\documentclass[runningheads]{llncs}
\usepackage[T1]{fontenc}
\usepackage{graphicx}
\usepackage{amsmath}
\usepackage{amssymb}
\usepackage{booktabs}
\usepackage{amssymb}
\usepackage{float}
\usepackage{subcaption}
\usepackage{subcaption}
\usepackage{multirow}

\begin{document}
\title{Ego4OOD: Rethinking Egocentric Video Domain Generalization via Covariate Shift Scoring}
\titlerunning{Ego4OOD: Rethinking Egocentric Video Domain Generalization}
\author{Zahra Vaseqi\orcidID{0000-0003-0098-8648} \and
James Clark\orcidID{0000-0002-4512-6171}}
\authorrunning{Z. Vaseqi et al.}

\institute{McGill University, Montreal QC H3A 0G4, Canada \\
\email{zahra.vaseqi@mail.mcgill.ca} \\
\email{james.clark1@mcgill.ca}}
\maketitle              
\begin{abstract}
Egocentric video action recognition under domain shifts remains challenging due to large intra-class spatio-temporal variability, long-tailed feature distributions, and strong correlations between actions and environments. Existing benchmarks for egocentric domain generalization often conflate covariate shifts with concept shifts, making it difficult to reliably evaluate a model’s ability to generalize across input distributions.
To address this limitation, we introduce Ego4OOD, a domain generalization benchmark derived from Ego4D that emphasizes measurable covariate diversity while reducing concept shift through semantically coherent, moment-level action categories. Ego4OOD spans eight geographically distinct domains and is accompanied by a clustering-based covariate shift metric that provides a quantitative proxy for domain difficulty.
We further leverage a one-vs-all binary training objective that decomposes multi-class action recognition into independent binary classification tasks. This formulation is particularly well-suited for covariate shift by reducing interference between visually similar classes under feature distribution shift.
Using this formulation, we show that a lightweight two-layer fully connected network achieves performance competitive with state-of-the-art egocentric domain generalization methods on both Argo1M and Ego4OOD, despite using fewer parameters and no additional modalities. Our empirical analysis demonstrates a clear relationship between measured covariate shift and recognition performance, highlighting the importance of controlled benchmarks and quantitative domain characterization for studying out-of-distribution generalization in egocentric video.

\end{abstract}

\section{Introduction}
Independent and identically distributed (i.i.d.) data is a key assumption in many research problems. This assumption is often violated across training and test data in practice.
Image and video classification models are prone to learn spurious correlations in visual content; this compromises their ability to generalize when a distributional shift between train and test sets is present. Such data biases are inherent in real-world computer vision datasets due to capture bias and geographical constraints \cite{dataset_bias} in data collection. When a strong correlation between two features in the training set exists, and the test set does not have the same type of correlation, the generalization performance suffers \cite{pretrain_aug_ds}. 

The issue with data distributional shifts has been widely studied in the context of synthetic shifts in image classification; however, it is unclear whether the synthetic robustness methods transfer to natural shifts \cite{HendrycksBMKWDD21,nat_ds_images,syn_vs_nat}.

While training on incrementally larger and more diverse datasets \cite{nat_ds_images} with data augmentation \cite{augmix,pretrain_aug_ds} have been shown to improve robustness in presence of distributional shifts; we still lack a comprehensive understanding of the underlying factors in model training that leads to performance discrepancies across domains. 

The notion of in-distribution in video understanding is inherently broad, encompassing a wide range of spatio-temporal variations observed during training. These include changes in motion dynamics, illumination (e.g., overcast or nighttime scenes), background clutter, and camera viewpoints that preserve the underlying semantic task. At the same time, shifts along these same factors—such as transitions between indoor and outdoor environments, extreme lighting conditions, or novel viewpoints—may constitute out-of-distribution data when they fall outside the support of the training distribution. As a result, out-of-distribution in videos is not defined by the presence of a specific factor, but by whether that factor, or its interaction with others over time, is sufficiently represented during training.

The image domain generalization benchmarks, where out-of-distribution shifts are typically defined through explicit domain splits targeting specific nuisance factors, such as image style \cite{PACS,office_home,domainnet}, color \cite{coloredmnist}, rotation \cite{rotatedmnist}, or background \cite{TerraIncognita}. In videos, the distinction between in-distribution and out-of-distribution data is often less clear. 
This issue is further amplified in egocentric video, where the data distribution is strongly conditioned on the camera wearer’s behavior and environment. Non-stationary camera motion, rapidly varying viewpoints, and task-driven interactions induce large intra-class spatio-temporal variability, while user-specific and long-tailed activity distributions imply that the training data covers only a sparse subset of possible factor combinations. Consequently, egocentric video substantially blurs the distinction between in-distribution variability and out-of-distribution data, making robust generalization particularly challenging. 

In this work, we investigate domain generalization in egocentric action recognition. We find that even a simple two-layer fully connected network can achieve performance on par with current state-of-the-art methods for domain generalization on egocentric action recognition. To support further research, we introduce Ego4OOD, a domain generalization benchmark built on the existing Ego4D dataset that reorganizes and categorizes videos to evaluate out-of-distribution generalization. Furthermore, we propose a covariate shift metric that provides a quantitative score for each domain, enabling systematic analysis of how visual features vary across domains. Our qualitative observations of visual differences across domains are consistent with the quantitative covariate shift scores, highlighting a strong correlation between perceived domain variation and the metric.

The contributions of this work are threefold: (i) we highlight the effectiveness of a simple network architecture for domain generalization in egocentric action recognition, (ii) we introduce Ego4OOD, a benchmark for systematic evaluation of out-of-distribution generalization in egocentric videos, and (iii) we provide an accompanying covariate shift metric and analysis that captures inter-domain feature variations. In the following sections, we first review related work on domain generalization and egocentric video analysis, then describe the construction of the Ego4OOD benchmark and our experimental setup, and finally present results and analysis that illustrate the challenges and insights for generalization across diverse video domains.

\section{Related Work}
\subsection{Egocentric Action Recognition}
In traditional third-person vision, a fixed camera passively observes the scene, whereas in egocentric vision the camera is worn by the actor, recording firsthand interactions with the surrounding environment. Research in egocentric vision has grown rapidly in recent years, driven by the release of large-scale benchmarks such as Epic-Kitchens \cite{epickitchens}, Ego4D \cite{grauman_ego4d_nodate}, Charades-Ego \cite{charadesego}. In egocentric videos, the environment often occupies a large portion of the field of view, and rapidly changing viewpoints make it challenging to recognize actions based solely on scene context or motion trajectories, which are more informative and stable in third-person videos \cite{wang2013action,zhao2018trajectory}. This highlights the importance of characterizing actions independently of the environment and generalizing to new domains \cite{peirone2025egocentric}. Ego-Topo \cite{nagarajan2020ego} introduces the notion of \textit{affordances} in egocentric vision, modeling actions as dependent on the environment and objects present—for example, a kitchen sink affording washing. EgoZAR \cite{peirone2025egocentric} highlights a key failure mode of this approach by modeling environmental affordances as \textit{activity-centric zones} derived from co-occurrence biases. While domain-specific affordances can improve performance when actions are evaluated in previously seen environments, it negatively impacts generalization to unseen environments.
EgoZAR \cite{peirone2025egocentric} suggests an adversarial classifier to extract domain-agnostic zone psuedo labels from the motion features. On the other hand, Plizzari et al. propose Cross-Instance Reconstruction (CIR) \cite{plizzari2023can}, which represents an action as a weighted combination of video instances drawn from other domains. CIR encourages domain-agnostic representations by reconstructing each video at the feature level using a support set of samples from other domains within the same batch. The reconstructed representation is then aligned with the corresponding textual narration via a contrastive video–text objective. By learning to reconstruct the same action from samples from other domains, CIR promotes representations that generalize beyond environment-specific biases. We benchmark our two-layer fully connected architecture against EgoZAR and CIR, demonstrating that it achieves comparable performance to these specialized methods, despite not leveraging text narrations, and can outperform them by a substantial margin in some cases with strong domain shift.

\subsection{Distribution Shift in Egocentric Video Datasets}
Distribution shift is commonly formalized as a mismatch between the joint distribution of the source training data $p_S(x,y)$ and the target test data $p_T(x,y)$, and is typically decomposed into covariate shift, concept shift, and prior shift \cite{farahani_brief_2020,quionero2009dataset}. \textbf{Covariate shift} occurs when the marginal distribution of input features $p(x)$ differs between source and target domains while the conditional distribution $p(y \mid x)$ remains unchanged; variations in lighting, background, or viewpoint are common examples. \textbf{Concept shift} arises when the conditional distribution $p(y \mid x)$ changes across domains. Ambiguity in class boundaries \cite{tsipras2020imagenet} is a fundamental source of concept shift. For instance, in ARGO1M \cite{plizzari2023can}, the largest domain generalization dataset for egocentric action recognition, action labels such as \emph{take}, \emph{lift}, \emph{carry}, and \emph{bring} correspond to visually similar motions but differ in task semantics or annotator interpretation.
Previous work on domain generalization benchmarks highlights that effective evaluation requires both sufficient distributional diversity and consistent category knowledge across domains \cite{zhang2023nico++}. Consistent category knowledge limits concept shift, while feature diversity across domains creates the generalization challenge by inducing covariate shifts. Zhang et al. \cite{zhang2023nico++} observe that benchmarks combining significant covariate shift with limited concept shift provide the most informative assessment of a model’s ability to generalize across domains.
Argo1M, by contrast, consists of 60 verb-based action classes with significant semantic overlap, making it susceptible to concept shift. To address this, we introduce Ego4OOD, a benchmark derived from Ego4D that emphasizes measurable covariate diversity while limiting concept shift. 
 
\textbf{Prior shift} occurs when the marginal distribution of labels p(y) differs between source and target domains. For example, in Ego4D \cite{grauman_ego4d_nodate}, videos collected by IIIT Hyderabad (iiith) in India contain a high proportion of food-preparation scenarios. In our geographically based domain generalization benchmark, Ego4OOD, this creates label imbalance: using a leave-one-location-out protocol--training on all other locations and testing on iiith—the test set exhibits a different label distribution, illustrating prior shift.

\begin{table}[t]
\centering
\scriptsize
\setlength{\tabcolsep}{3pt}
\renewcommand{\arraystretch}{1.1}
\begin{tabular}{@{}lp{2cm}p{5cm}@{}}
\toprule
\textbf{Source vs. Target Distributions} & \textbf{Shift Type} & \textbf{} \\ 
\midrule
\begin{tabular}[c]{@{}l@{}}$p_S(x) = p_T(x)$ \\ $p_S(y) = p_T(y)$ \\ $p_S(y|x) = p_T(y|x)$ \end{tabular} 
& \textbf{No Shift} 
& i.i.d. assumption holds.\\\\

$p_S(x) \neq p_T(x)$ & \textbf{Covariate Shift} & Same class categories but different input setting; e.g., indoor vs. outdoor. \\\\

$p_S(y|x) \neq p_T(y|x)$ & \textbf{Concept Shift} & Similar inputs map to different class categories; e.g., visually similar hand motions labeled as \emph{carry} or \emph{bring} based on annotator choice. \\\\

$p_S(y) \neq p_T(y)$ & \textbf{Prior Shift} & Class frequencies differ. \\
\bottomrule
\end{tabular}
\caption{Types of distribution shifts.}
\label{tab:domain_shifts}
\end{table}

\section{Ego4OOD Benchmark}
We introduce Ego4OOD, a domain generalization benchmark constructed from moment-level annotations in the Ego4D dataset \cite{grauman_ego4d_nodate}. Ego4D moments provide fine-grained, temporally localized descriptions of human actions, covering over 203 atomic activities.
We group the 203 moment classes into 9 semantically coherent categories, reducing ambiguity and limiting concept shift while preserving covariate diversity across domains. Table~\ref{tab:ego4ood_moments} shows example moment annotations grouped into 9 categories with coherent conceptual boundaries. 

While our categorization is conceptually similar to the video-level scenario annotations provided in Ego4D, those annotations are assigned to full-length videos, which can span several hours. A single video may be labeled with multiple scenarios without specifying the temporal extent of each, making the annotations imprecise as context often changes throughout the video.
To address this, we leverage the granular moment-level annotations in Ego4D to define our category mapping, providing more precise semantic boundaries and reducing ambiguity.
The full set of moment-to-category mappings is provided in Appendix~1. Ego4OOD comprises 20,096 video clips spanning eight distinct regions: the UK, India, Italy, Saudi Arabia, FRL (Facebook Reality Labs office), US-CMU, US-Minnesota, and Japan. 

\begin{figure*}[t]
    \centering
    \setlength{\fboxsep}{0pt} 

    \begin{minipage}{\textwidth}
        \textbf{(a)}\hspace{2mm} 
        \includegraphics[width=0.15\textwidth]{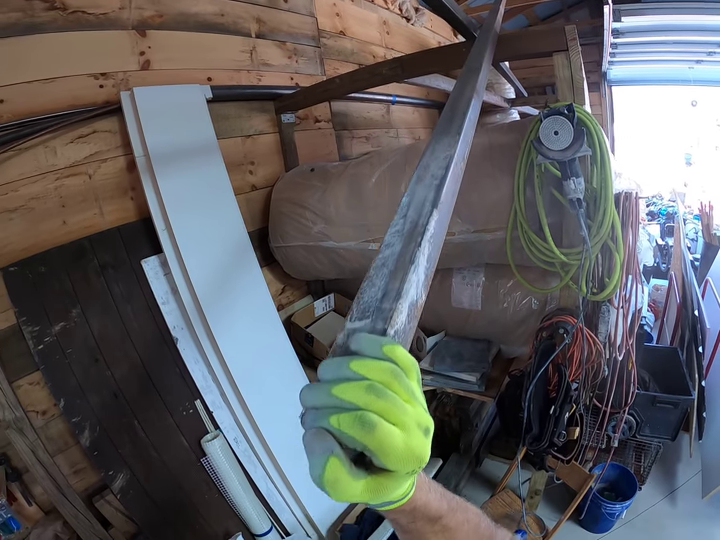}
        \includegraphics[width=0.15\textwidth]{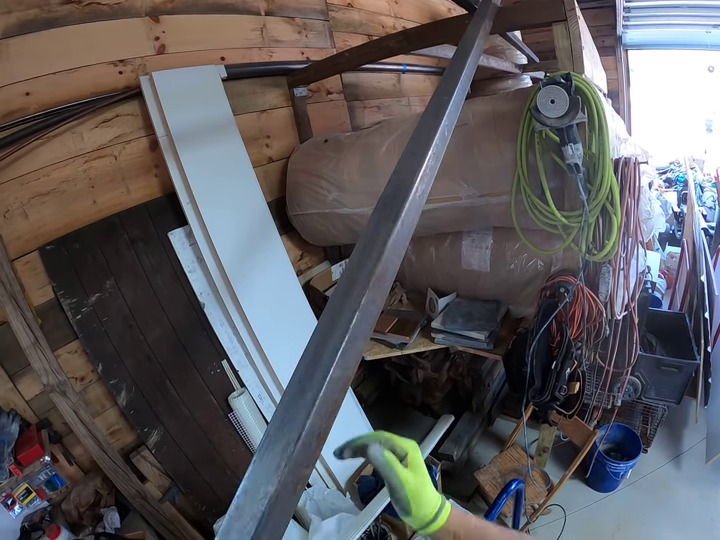}
        \includegraphics[width=0.15\textwidth]{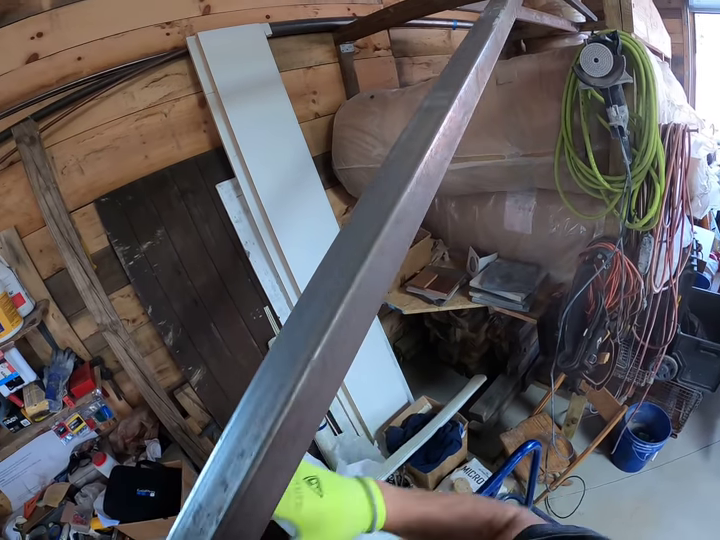}
        \includegraphics[width=0.15\textwidth]{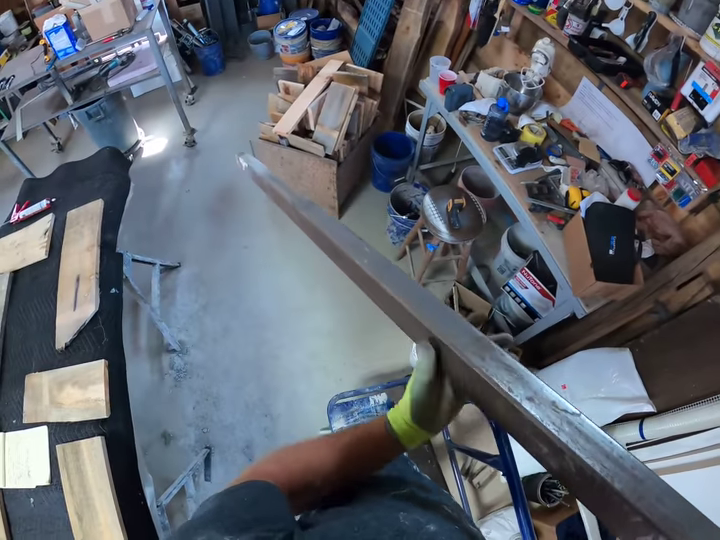}
        \includegraphics[width=0.15\textwidth]{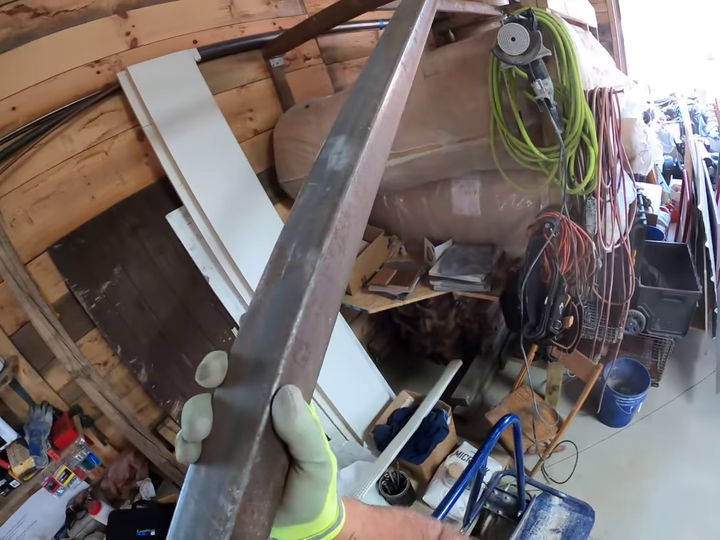}
        \includegraphics[width=0.15\textwidth]{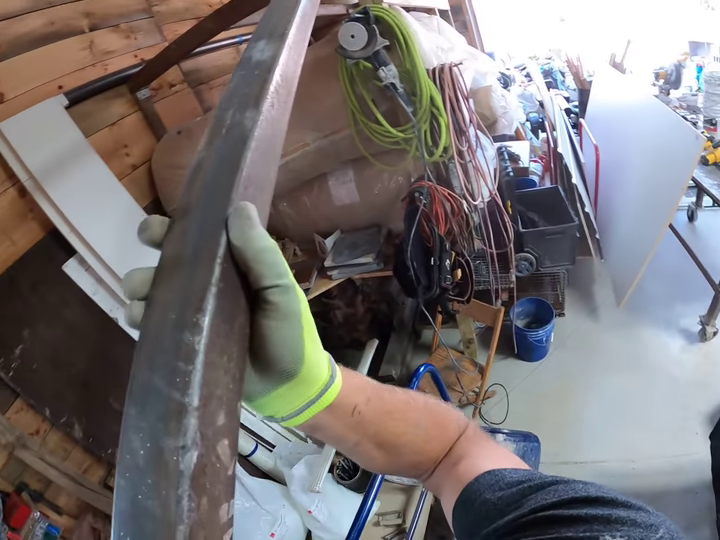}
    \end{minipage}

    \vspace{1mm} 

    \begin{minipage}{\textwidth}
        \textbf{(b)}\hspace{2mm}
        \includegraphics[width=0.15\textwidth]{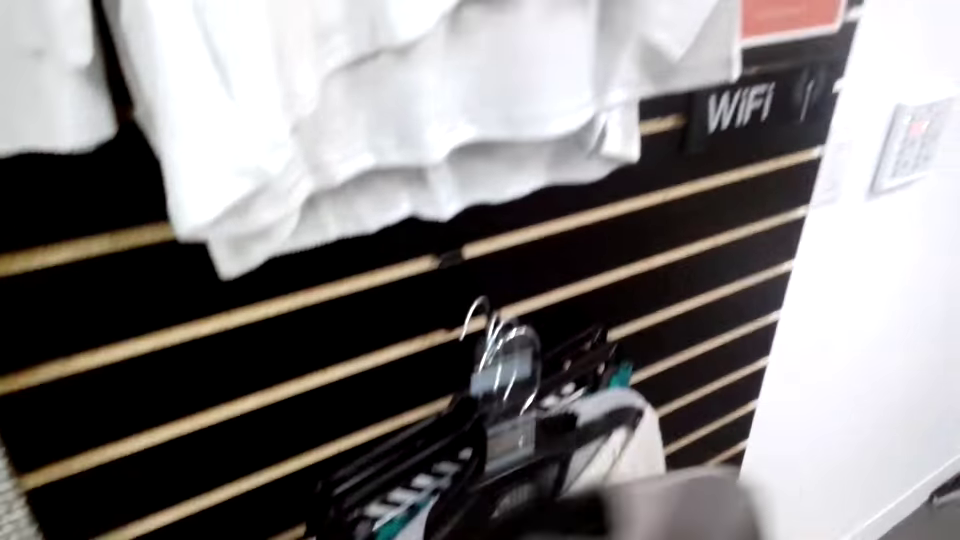}
        \includegraphics[width=0.15\textwidth]{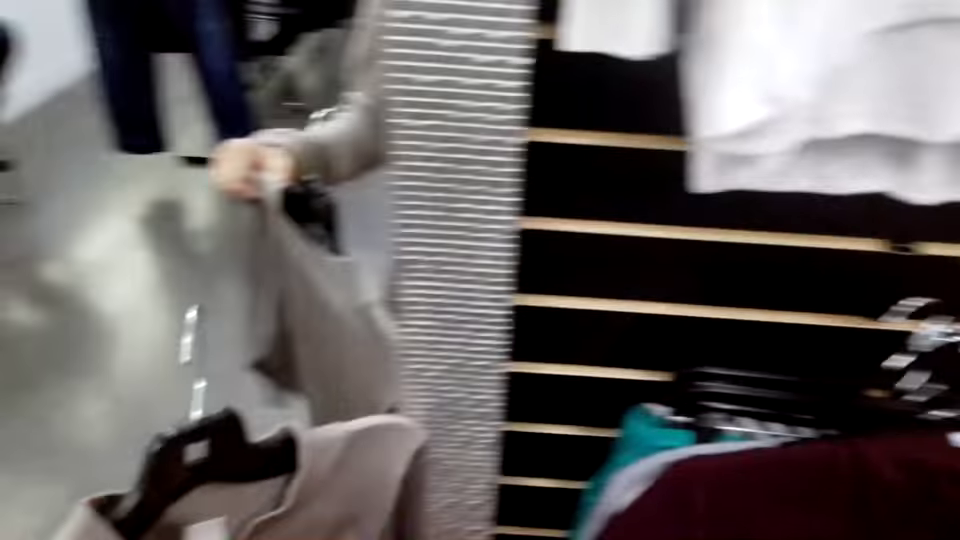}        \includegraphics[width=0.15\textwidth]{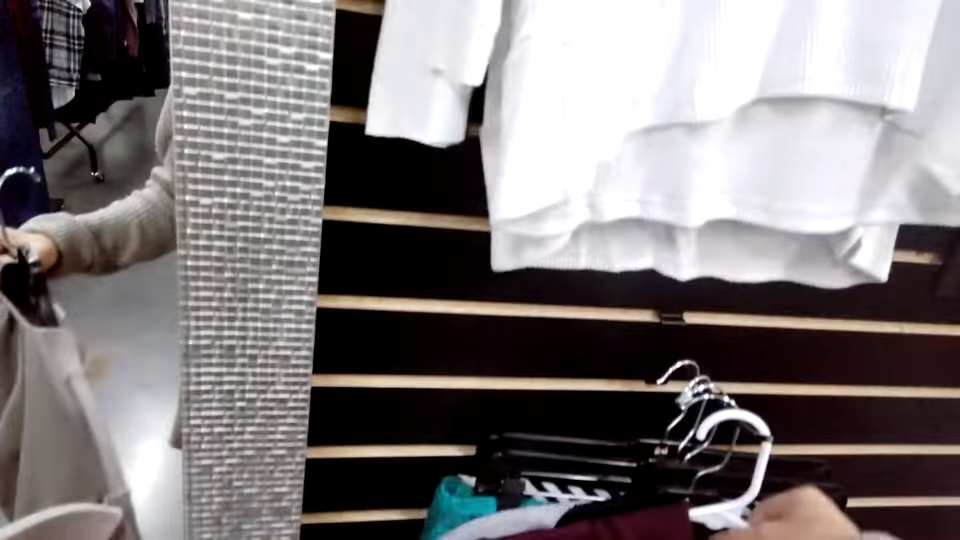}
        \includegraphics[width=0.15\textwidth]{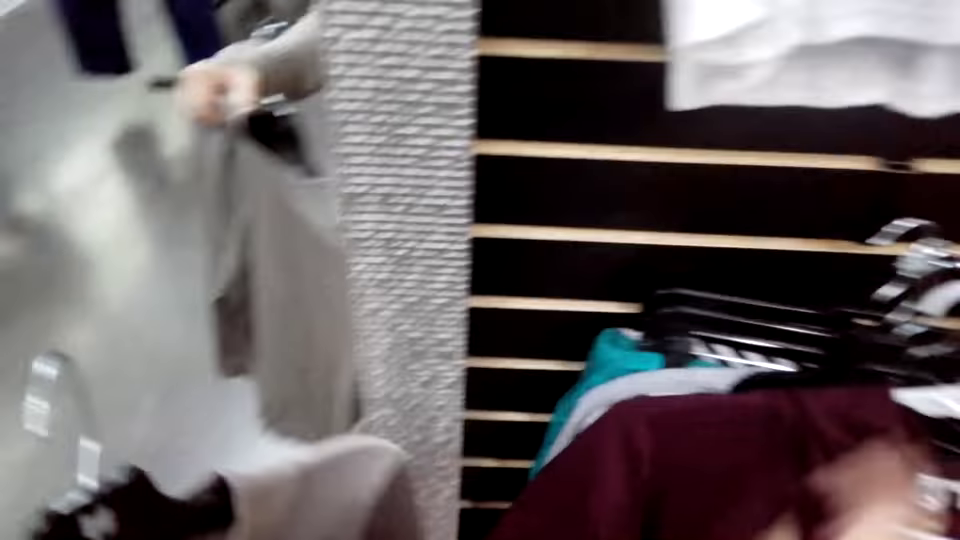}
        \includegraphics[width=0.15\textwidth]{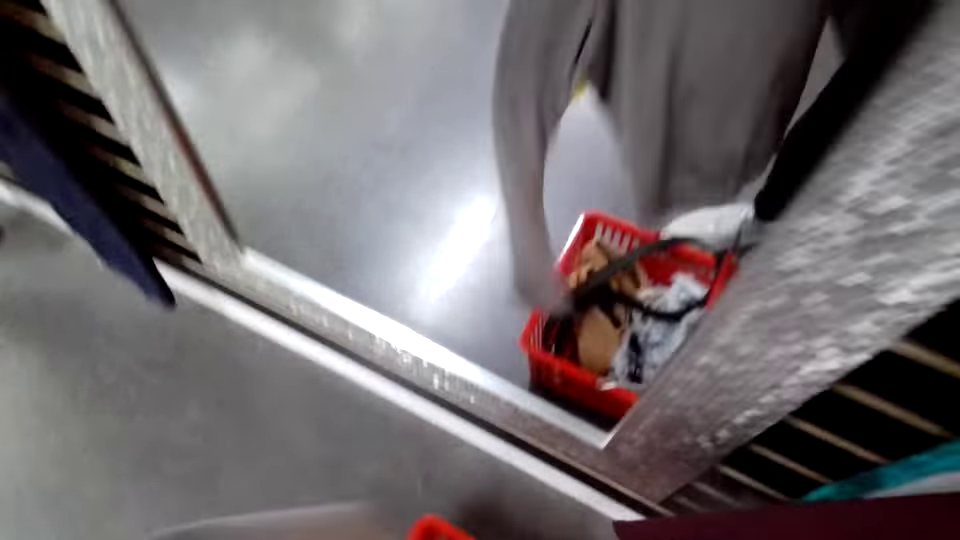}
        \includegraphics[width=0.15\textwidth]{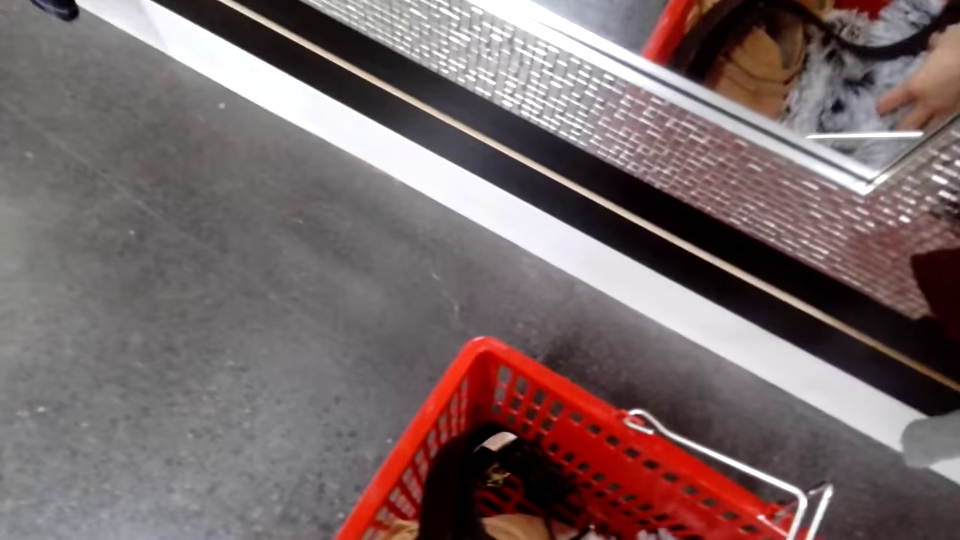}
    \end{minipage}
    
    \begin{minipage}{\textwidth}
        \textbf{(c)}\hspace{2mm}
        \includegraphics[width=0.15\textwidth]{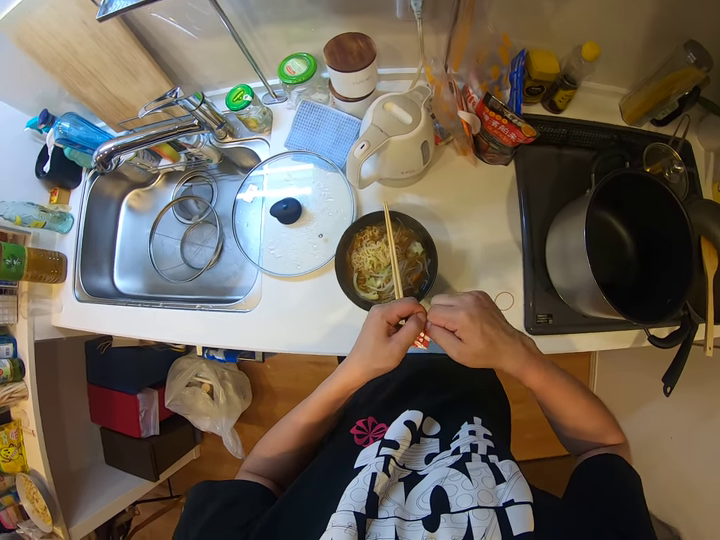}
        \includegraphics[width=0.15\textwidth]{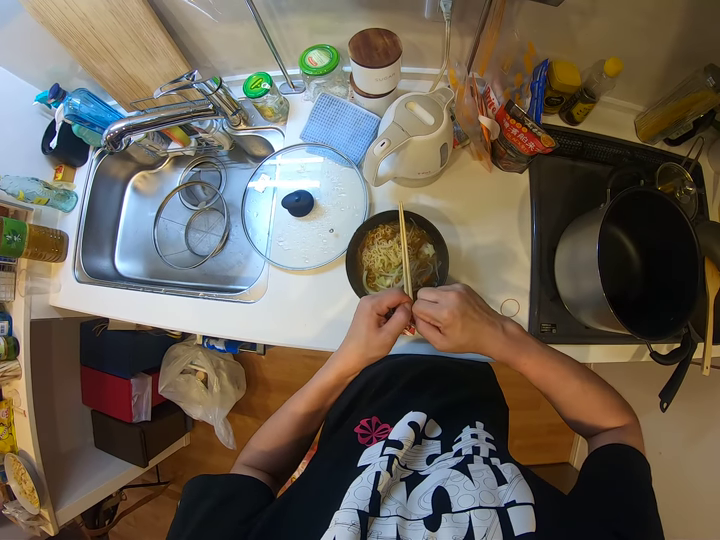}
        \includegraphics[width=0.15\textwidth]{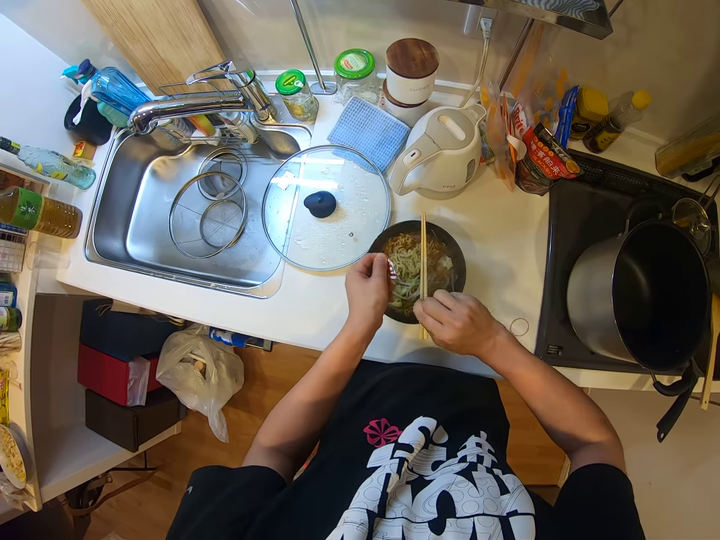}
        \includegraphics[width=0.15\textwidth]{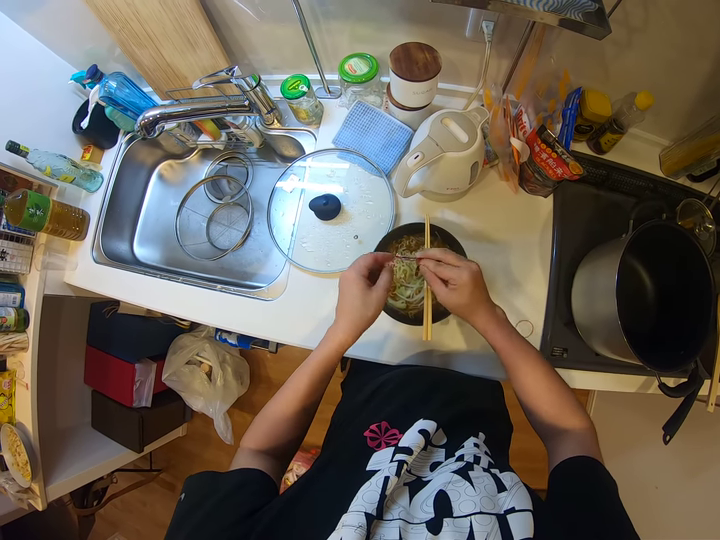}
        \includegraphics[width=0.15\textwidth]{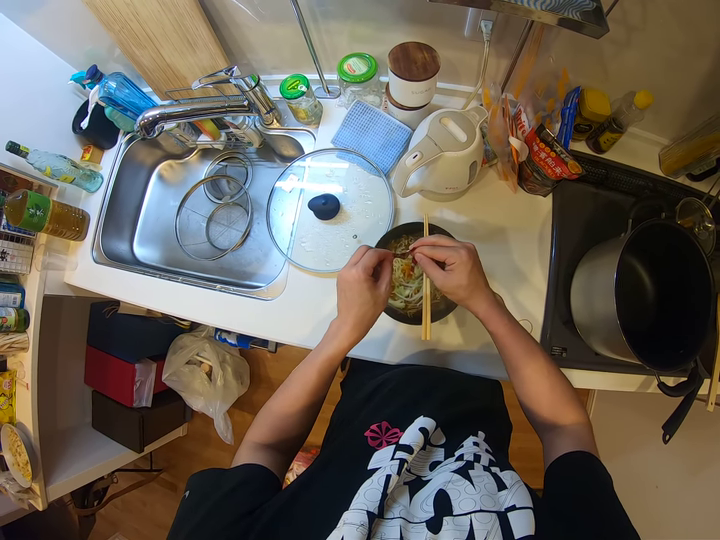}
        \includegraphics[width=0.15\textwidth]{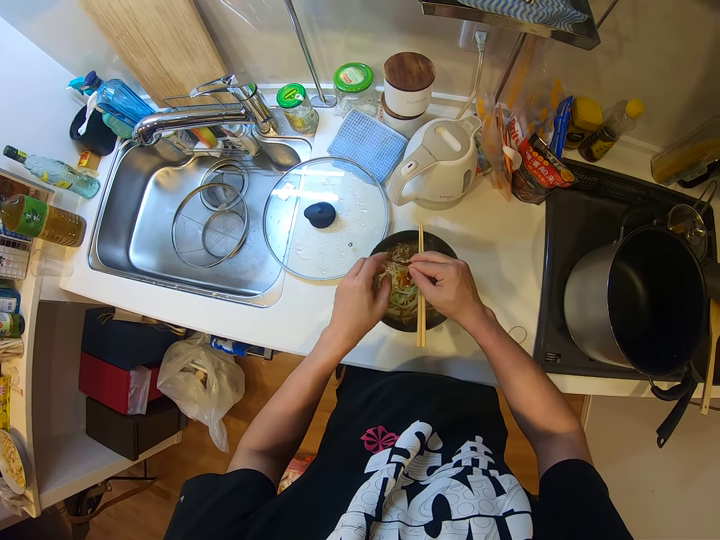}
    \end{minipage}

    \vspace{1mm}

    \begin{minipage}{\textwidth}
        \textbf{(d)}\hspace{2mm}
        \includegraphics[width=0.15\textwidth]{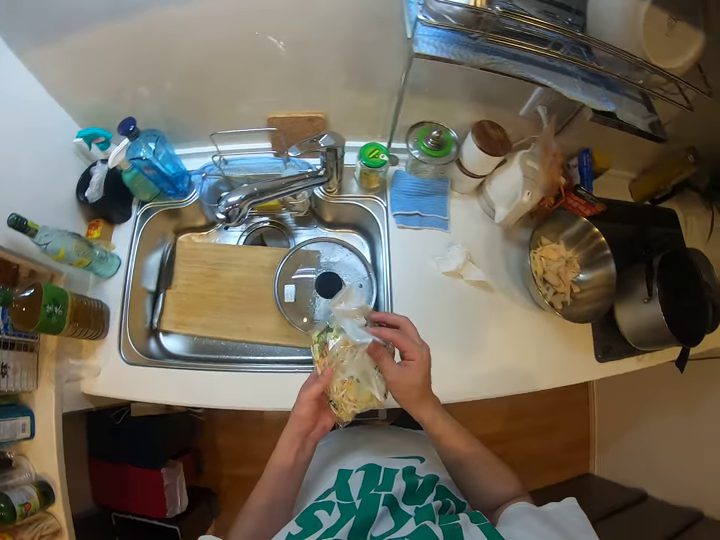}
        \includegraphics[width=0.15\textwidth]{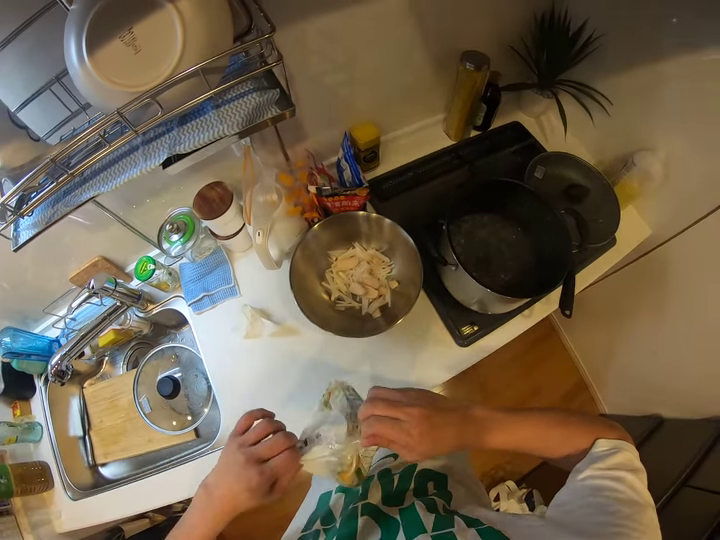}
        \includegraphics[width=0.15\textwidth]{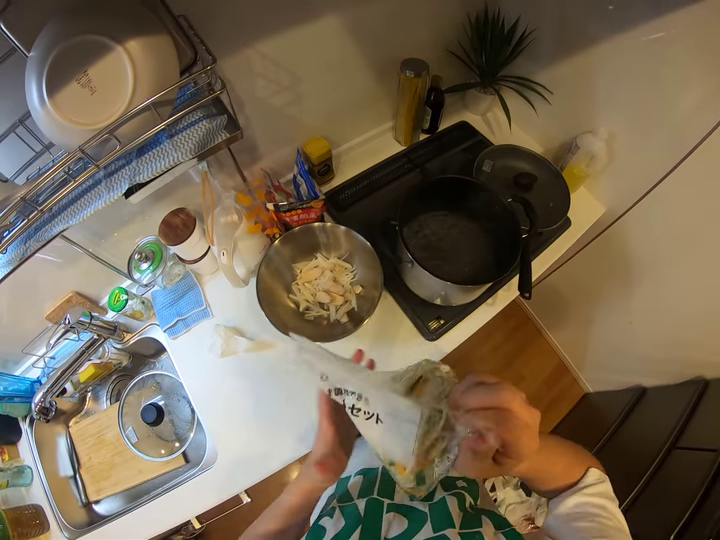}
        \includegraphics[width=0.15\textwidth]{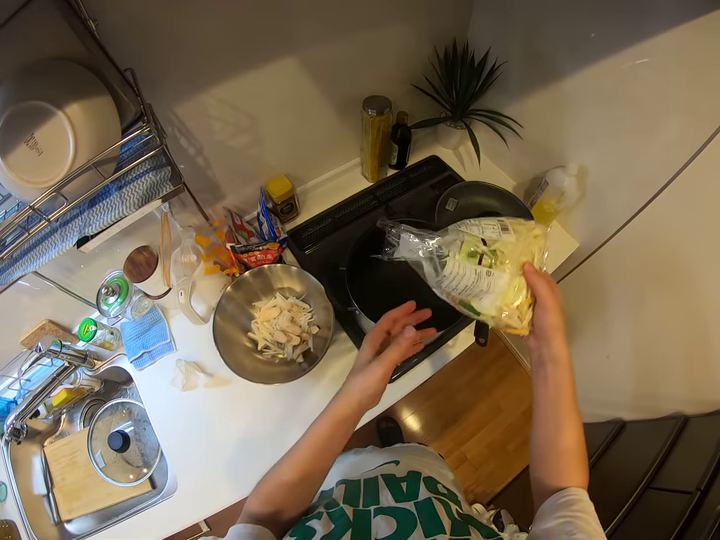}
        \includegraphics[width=0.15\textwidth]{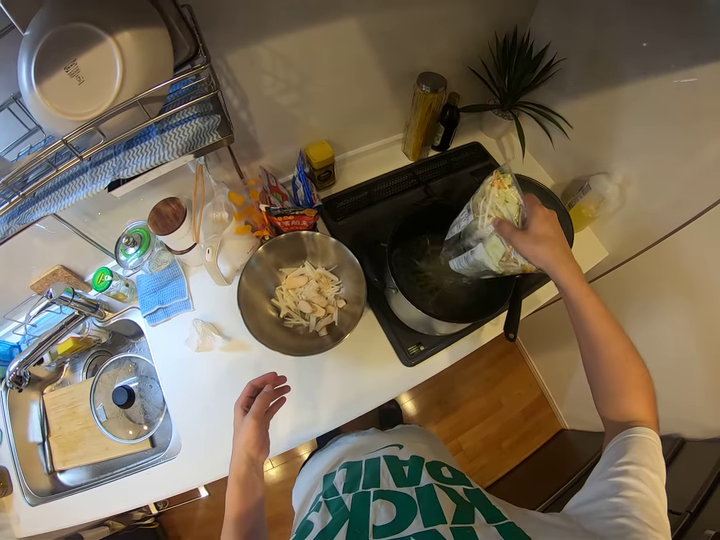}
        \includegraphics[width=0.15\textwidth]{frames/open_utokyo_d59f/frames_0010.png}
    \end{minipage}

    \vspace{1mm}
    \begin{minipage}{\textwidth}
        \textbf{(e)}\hspace{2mm}
        \includegraphics[width=0.15\textwidth]{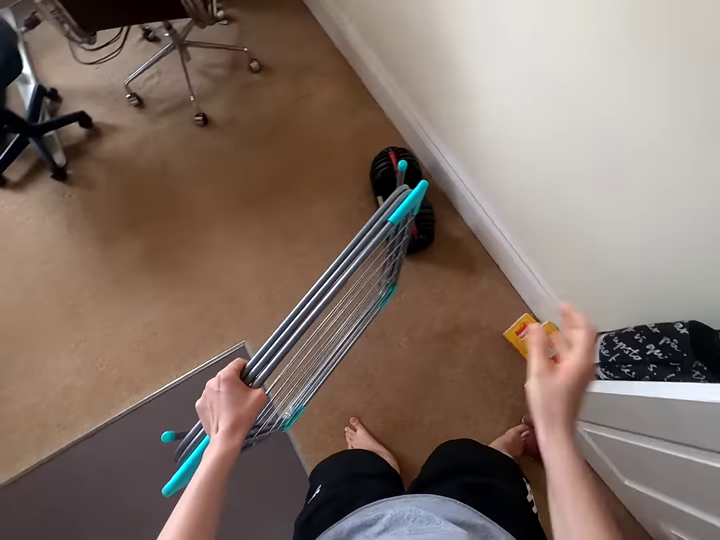}
        \includegraphics[width=0.15\textwidth]{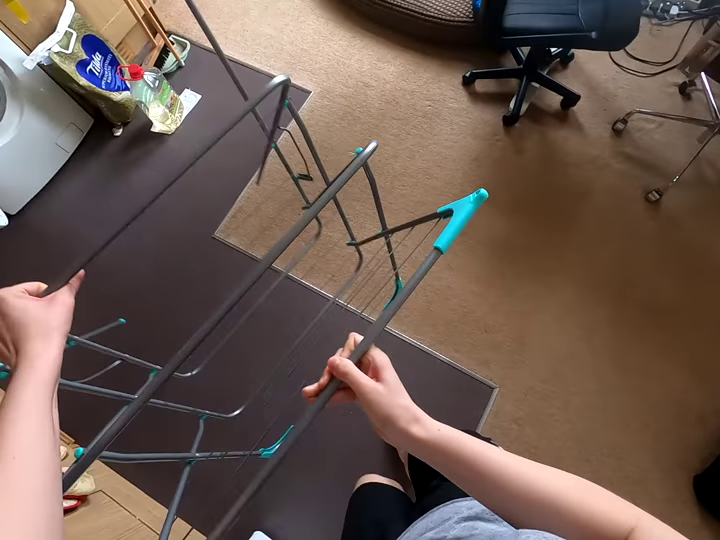}
        \includegraphics[width=0.15\textwidth]{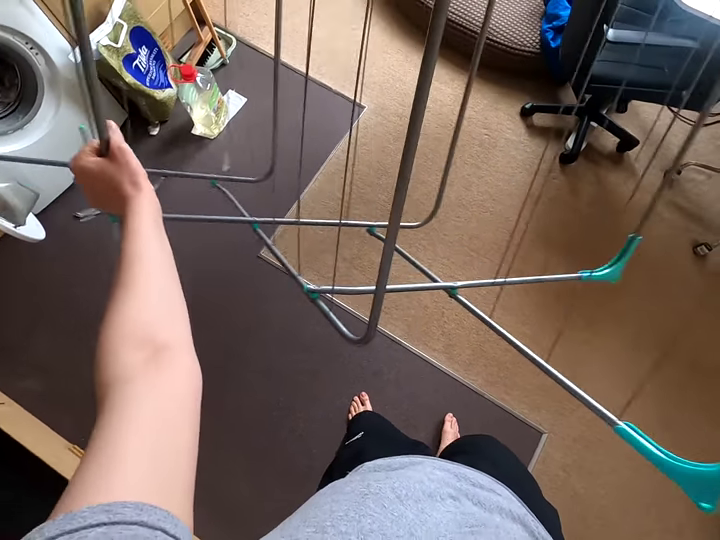}
        \includegraphics[width=0.15\textwidth]{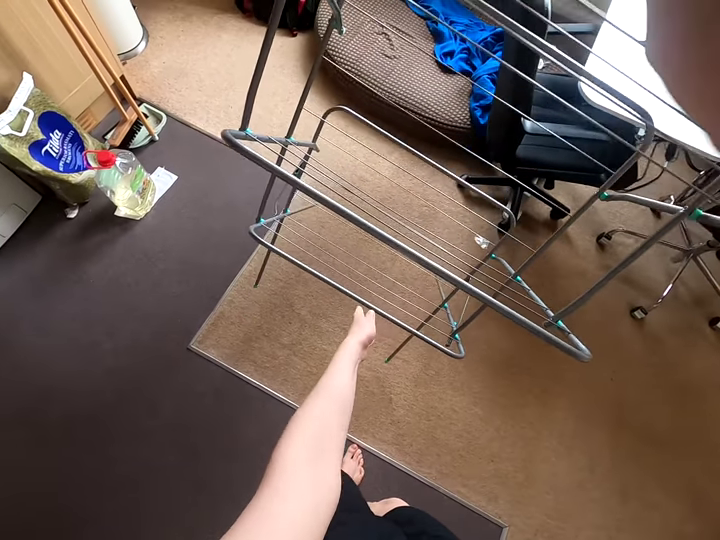}
        \includegraphics[width=0.15\textwidth]{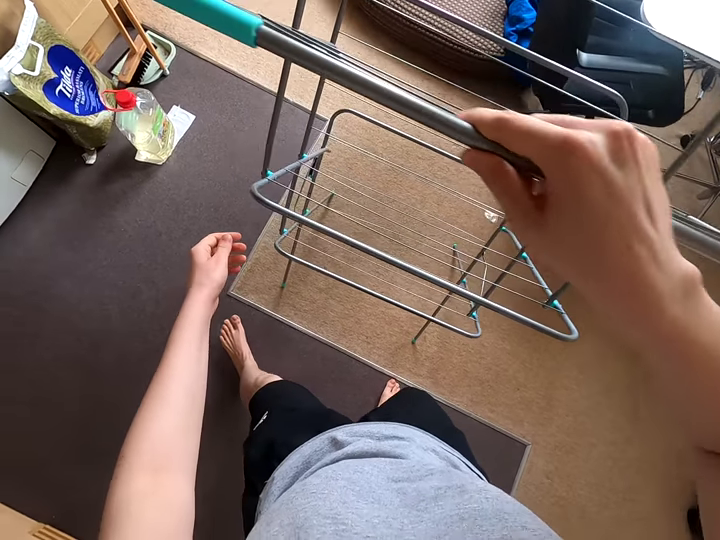}
        \includegraphics[width=0.15\textwidth]{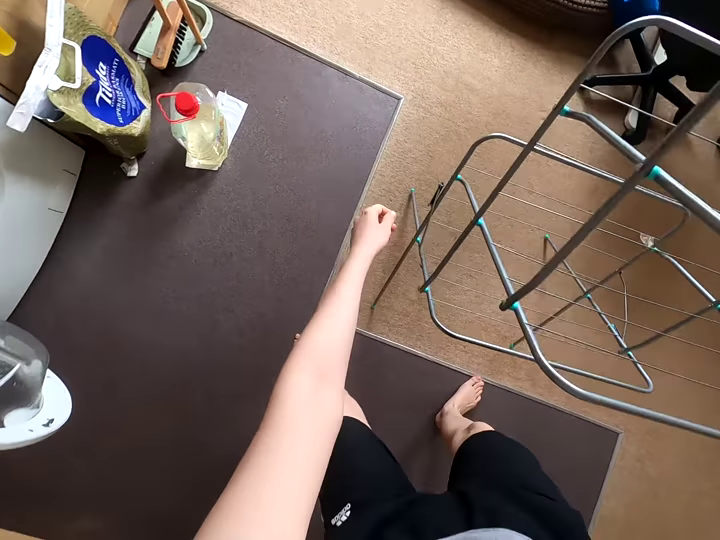}
    \end{minipage}

    \vspace{1mm}

    \caption{Example frames from five video clips, each row corresponding to a single clip as annotated in the Argo1M benchmark: (a) \textit{bring}, (b) \textit{pat}, (c) \textit{tear}, (d) \textit{open}, (e) \textit{bring}. These examples illustrate annotation ambiguities and co-occurring actions that can lead to misclassification, highlighting key failure modes of the Argo1M benchmark.}
    
    \label{fig:video_frames}
\end{figure*}

In Figure~\ref{fig:video_frames}, we illustrate representative examples from the Argo1M dataset that highlight how ambiguities in class definitions make it difficult to categorize actions into mutually exclusive classes. In Fig.~1(a), an action described as ``brings down a metal iron'' is annotated as \textit{bring}, even though both \textit{bring} and \textit{lower} are distinct classes in Argo1M, and the observed motion is equally consistent with \textit{lower}.
In Fig.~\ref{fig:video_frames} (b), the subject reaches toward a shirt on a hanger while, \emph{as seen via the mirror}, the other hand simultaneously lowers a different clothing item into a basket. Despite this, the clip is annotated as \textit{pat}, even though no patting action occurs during the annotated temporal window. Only later in the clip does the subject \textit{pat} a separate clothing item already inside the basket. This misalignment between the annotation and the visual content renders the clip {fundamentally ambiguous}, making it virtually impossible for a model to assign a correct and mutually exclusive action label based solely on the visual input.
In Fig.~\ref{fig:video_frames}(b), an action of tearing a sachet is labeled as \textit{tear}, though it could equally be interpreted as \textit{open} depending on the annotator’s preference. This inconsistency is illustrated in row (c), where tearing another sachet is labeled as \textit{open} rather than \textit{tear}.
Finally, in Fig.~\ref{fig:video_frames}(e), the actor brings out a hanger and opens it in the room, making the \textit{bring}, \textit{open}, and \textit{carry} categories all simultaneously valid. Because these actions co-occur, the annotation is inherently ambiguous, yet evaluation still requires selecting a single label.

While some of these issues stem from the inherent ambiguity of language, the category design in Argo1M makes it highly prone to misclassification, with state-of-the-art performance reaching only 24.92\% (random chance: 11.58\%). In contrast, Ego4OOD, despite a much smaller training set, benefits from semantically clear categories and high-quality moment-level annotations, achieving an average accuracy of 55.68\% (random chance: 13.47\%) with the same methods. This demonstrates that even with fewer training examples, well-defined categories substantially reduce ambiguity and improve model performance.

We design {Ego4OOD} to address the limitations of existing benchmarks such as Argo1M \cite{grauman_ego4d_nodate}.
While Argo1M contains roughly 1 million clips, Ego4OOD focuses on a carefully curated set of 20,096 clips, with a median test split of 1,152 clips compared to 16,916 in Argo1M, emphasizing quality over quantity. 
By selecting unambiguous, semantically coherent categories with clearly defined concept boundaries, Ego4OOD substantially reduces concept shift.  
Furthermore, the high-quality moment-level annotations ensure precise temporal grounding, supporting accurate evaluation of model performance on egocentric video tasks.
Ego4OOD will be made publicly available to facilitate reproducibility and future research.

\begin{table}[H]
\centering
\caption{Example Ego4D moment annotations mapped to Ego4OOD categories.}
\scriptsize
\setlength{\tabcolsep}{4pt}
\renewcommand{\arraystretch}{1.15}
\begin{tabular}{p{3cm} p{8.5cm}}
\toprule
\textbf{Category} & \textbf{Moments} \\
\midrule
Food Preparation &
assemble a sandwich, wash vegetables or fruit, whisk eggs in a bowl \\

House Chores &
make the bed, organize books on a shelf, try on clothing items \\

Gardening &
remove weeds, trim hedges or branches, apply fertilizer with a sprayer \\

Construction &
use a hammer to fix a nail, cut wood pieces, weld a metal item \\

Leisure &
walk the dog, dance to music, do exercise \\

Electronics Use &
use a phone, use a laptop computer, take photos or record videos \\

Personal Care &
wash hands, clean hands with sanitizer \\
\bottomrule
\end{tabular}
\label{tab:ego4ood_moments}
\end{table}

\subsubsection{Ego4OOD Splits.}
\begin{figure}[t]
  \centering
  \begin{subfigure}{0.48\linewidth}
    \includegraphics[width=\linewidth]{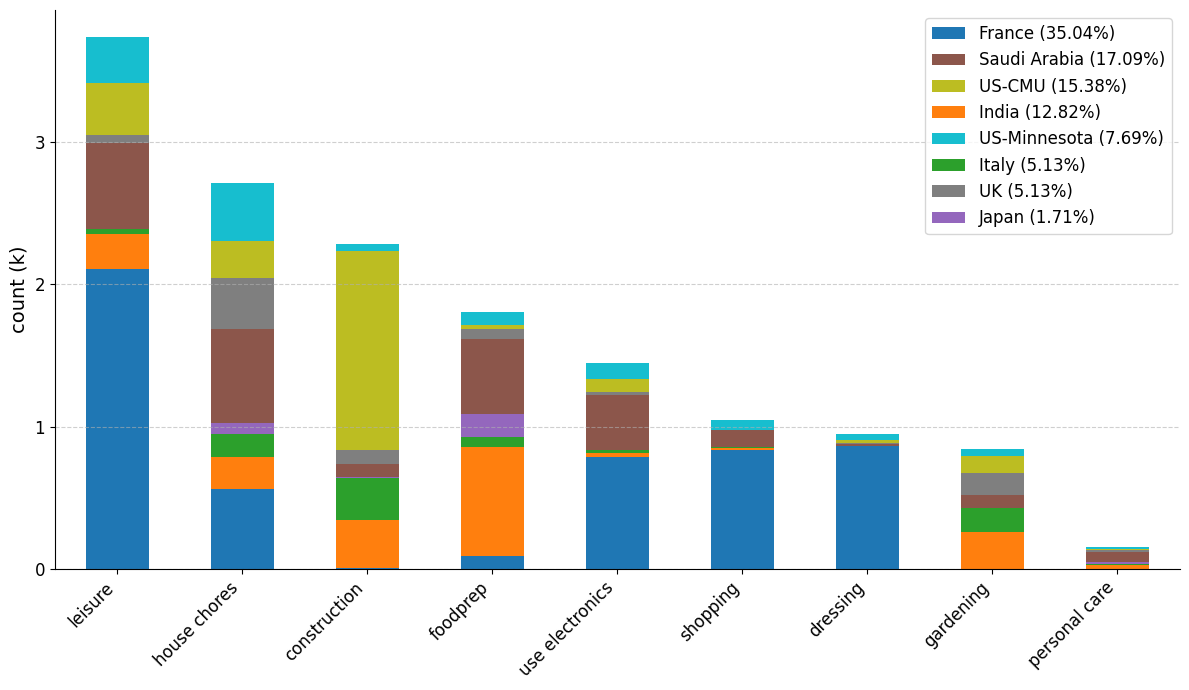}
  \end{subfigure}
  \hfill
  \begin{subfigure}{0.48\linewidth}
    \includegraphics[width=\linewidth]{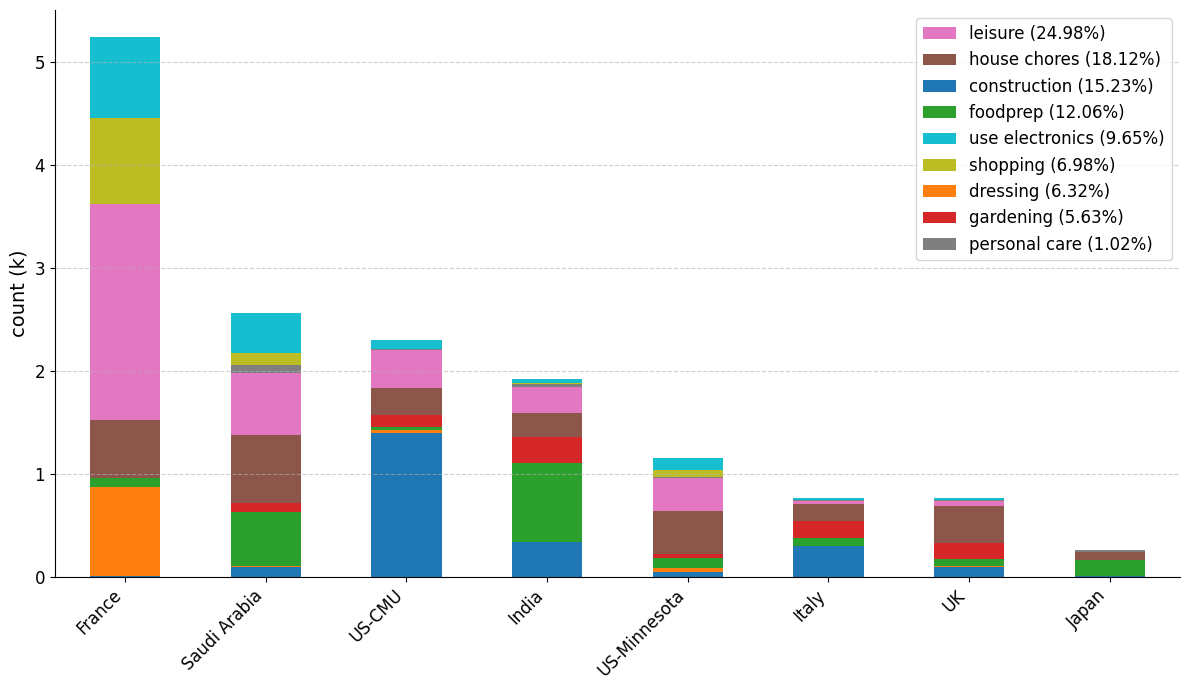}
  \end{subfigure}
  \caption{Number of video clips in Ego4OOD aggregated by (left) \textit{category} and (right) \textit{domain}.}
  \label{fig:ego4ood-clip-distribution}
\end{figure}

We use a leave-one-domain-out protocol to construct eight benchmarks. The number of samples in the train, validation, and test splits is summarized in Table~\ref{tab:domain_counts}. Furthermore, Fig.~\ref{fig:ego4ood-clip-distribution} illustrates the distribution of video clips across categories and domains in Ego4OOD, providing insights into its composition.

\begin{table}[t]
\centering
\caption{Ego4OOD benchmark sample counts per domain split (train/val/test).}
\label{tab:domain_counts}
\small
\setlength{\tabcolsep}{12pt} 
\renewcommand{\arraystretch}{1.2} 
\begin{tabular}{lccc}
\toprule
\textbf{Domain} & \textbf{Train} & \textbf{Val} & \textbf{Test} \\
\midrule
India         & 13,440 & 4,096 & 1,920 \\
FRL        & 10,112 & 3,072 & 5,248 \\
UK            & 14,592 & 4,736 & 768   \\
US-Minnesota  & 14,208 & 4,480 & 1,152 \\
US-CMU        & 13,056 & 4,224 & 2,304 \\
Saudi Arabia  & 12,800 & 4,096 & 2,560 \\
Italy         & 14,592 & 4,480 & 768   \\
Japan         & 15,104 & 4,736 & 256   \\
\bottomrule
\end{tabular}
\end{table}

\subsubsection{Analyzing Covariate Shift.}
We define a metric to measure covariate shift across domains, providing a quantitative proxy for domain difficulty.
Let $\mathbb{X} \in \mathbb{R}^{N \times D}$ be the feature matrix of $N$ samples, each with $D$-dimensional features. We first cluster all features using \textit{kmeans} into $K$ clusters:

\begin{equation}
\{\mathbf{c}_k \in \mathbb{R}^{D}\}_{k=1}^{K} = {kmeans}(\mathcal{X}, K),
\end{equation}
where $\mathbf{c}_k$ denotes the $k$-th cluster centroid in the feature space.
Each sample is associated with a domain $d \in \mathcal{D}$ and a class $y \in \mathcal{Y}$. Depending on the grouping mode, prototypes are computed for three types of grouping: 
\textbf{Domain-Class} $(d, y)$; \textbf{Domain} $d$: and \textbf{Class} $y$.
For a given group $g$, let the set of indices of samples in this group be

\begin{equation}
\mathcal{I}_g = \{i : \text{sample } i \text{ belongs to group } g\}.
\end{equation}
For each feature ${x}_i$ in the group, we assign it to the nearest cluster centroid:

\begin{equation}
k_i = \arg\min_{k} \; \text{dist}({x}_i, \mathbf{c}_k),
\end{equation}
where $\text{dist}(\cdot, \cdot)$ is the Euclidean distance. The group prototype is then the mean of the assigned centroids:

\begin{equation}
\mathbf{p}_g = \frac{1}{|\mathcal{I}_g|} \sum_{i \in \mathcal{I}_g} \mathbf{c}_{k_i}.
\end{equation}
For each group prototype $\mathbf{p}_g$, we compute its distance to all other prototypes $\mathbf{p}_{g'}$, $g' \neq g$, to measure its out-of-distributionness:

\begin{equation}
\Delta_g = \{ \text{dist}(\mathbf{p}_g, \mathbf{p}_{g'}) \; | \; g' \neq g \}.
\end{equation}
We summarize these pairwise distances with:
\begin{equation}
\mu_g = \frac{1}{|\Delta_g|} \sum_{\delta \in \Delta_g} \delta,
\qquad
\sigma_g = \sqrt{ \frac{1}{|\Delta_g|} \sum_{\delta \in \Delta_g} (\delta - \mu_g)^2 },
\end{equation}
where $\mu_g$ is the mean distance from the group prototype $\boldsymbol{\mu}_g$ to other prototypes, and $\sigma_g$ captures the variability of these distances. 
Intuitively, $\mu_d$ captures the average separation of a domain from the rest of the data distribution, while the $2\sigma_d$ term accounts for variability in this separation across different subgroups and class compositions within the domain.
To quantify covariate shift at the domain level, we summarize the distances between a domain prototype and all other domain prototypes using the composite metric
$\Omega(g) = \mu_g + \tau\sigma_g$. We set $\tau = 2$ and refer to $\Omega(g)$ as the domain shift score. Table~\ref{tab:d_ood_scores_with_ood} presents domain shift scores across Ego4OOD domains.

\begin{figure*}[t]
    \centering
    \setlength{\fboxsep}{0pt} 
    \renewcommand{\arraystretch}{0.2} 

    \begin{tabular}{c l}
        \hline
        \multirow{3}{*}{\rotatebox[origin=c]{90}{India}} &
        \includegraphics[width=0.12\textwidth]{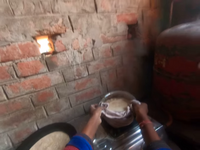}
        \includegraphics[width=0.12\textwidth]{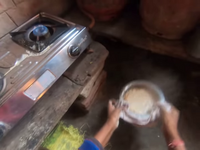}
        \includegraphics[width=0.12\textwidth]{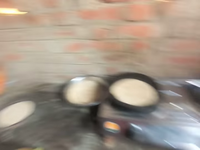}
        \includegraphics[width=0.12\textwidth]{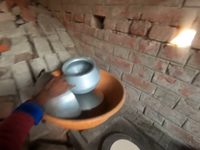}
        \includegraphics[width=0.12\textwidth]{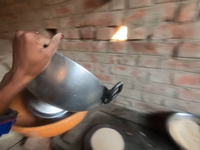}
        \includegraphics[width=0.12\textwidth]{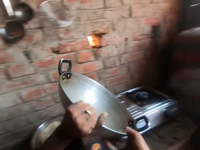} \\

        & 
        \includegraphics[width=0.12\textwidth]{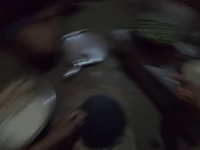}
        \includegraphics[width=0.12\textwidth]{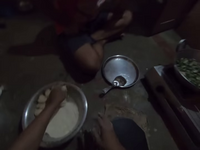}
        \includegraphics[width=0.12\textwidth]{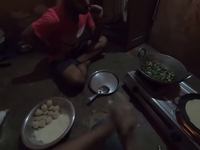}
        \includegraphics[width=0.12\textwidth]{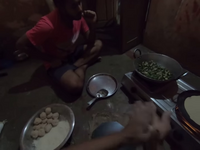}
        \includegraphics[width=0.12\textwidth]{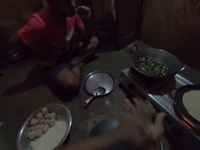}
        \includegraphics[width=0.12\textwidth]{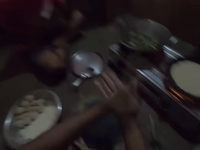} \\

        & 
        \includegraphics[width=0.12\textwidth]{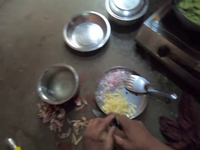}
        \includegraphics[width=0.12\textwidth]{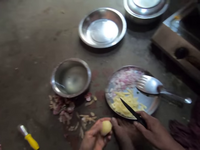}
        \includegraphics[width=0.12\textwidth]{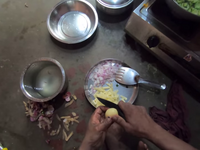}
        \includegraphics[width=0.12\textwidth]{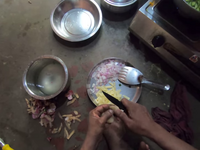}
        \includegraphics[width=0.12\textwidth]{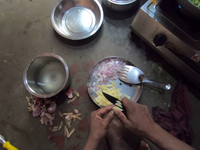}
        \includegraphics[width=0.12\textwidth]{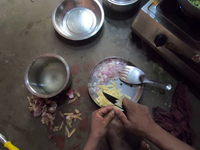} \\
        \hline

        \multirow{3}{*}{\rotatebox[origin=c]{90}{FRL}} &
        \includegraphics[width=0.12\textwidth]{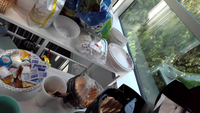}
        \includegraphics[width=0.12\textwidth]{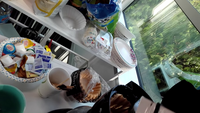}
        \includegraphics[width=0.12\textwidth]{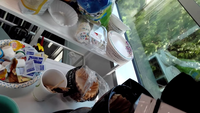}
        \includegraphics[width=0.12\textwidth]{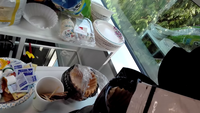}
        \includegraphics[width=0.12\textwidth]{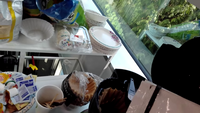}
        \includegraphics[width=0.12\textwidth]{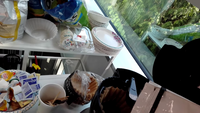} \\

        & 
        \includegraphics[width=0.12\textwidth]{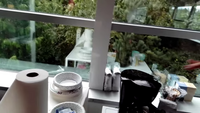}
        \includegraphics[width=0.12\textwidth]{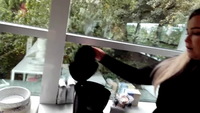}
        \includegraphics[width=0.12\textwidth]{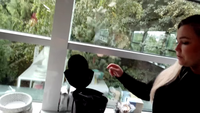}
        \includegraphics[width=0.12\textwidth]{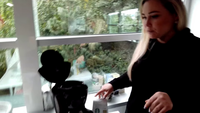}
        \includegraphics[width=0.12\textwidth]{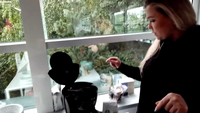}
        \includegraphics[width=0.12\textwidth]{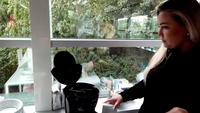} \\

        & 
        \includegraphics[width=0.12\textwidth]{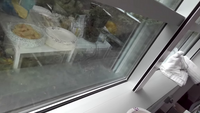}
        \includegraphics[width=0.12\textwidth]{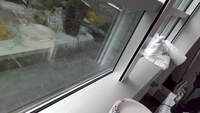}
        \includegraphics[width=0.12\textwidth]{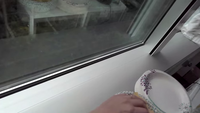}
        \includegraphics[width=0.12\textwidth]{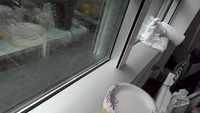}
        \includegraphics[width=0.12\textwidth]{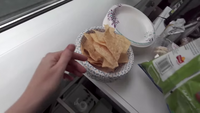}
        \includegraphics[width=0.12\textwidth]{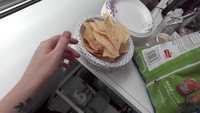} \\
        \hline

        \multirow{3}{*}{\rotatebox[origin=c]{90}{UK}} &
        \includegraphics[width=0.12\textwidth]{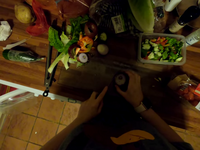}
        \includegraphics[width=0.12\textwidth]{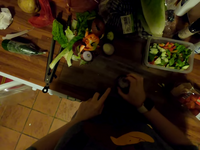}
        \includegraphics[width=0.12\textwidth]{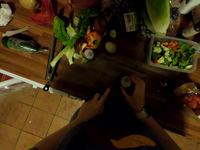}
        \includegraphics[width=0.12\textwidth]{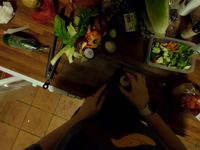}
        \includegraphics[width=0.12\textwidth]{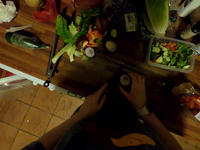}
        \includegraphics[width=0.12\textwidth]{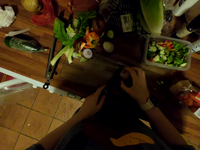} \\

        & 
        \includegraphics[width=0.12\textwidth]{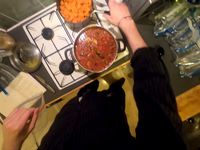}
        \includegraphics[width=0.12\textwidth]{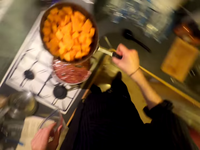}
        \includegraphics[width=0.12\textwidth]{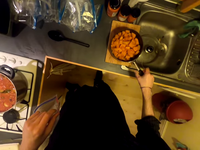}
        \includegraphics[width=0.12\textwidth]{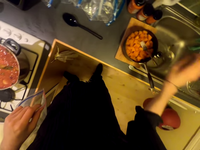}
        \includegraphics[width=0.12\textwidth]{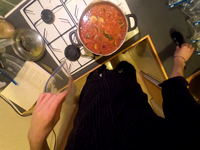}
        \includegraphics[width=0.12\textwidth]{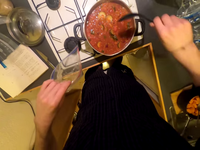} \\

        & 
        \includegraphics[width=0.12\textwidth]{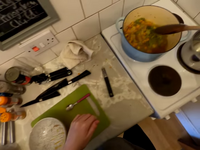}
        \includegraphics[width=0.12\textwidth]{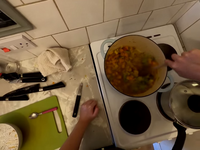}
        \includegraphics[width=0.12\textwidth]{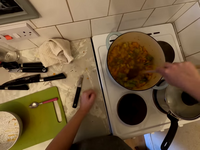}
        \includegraphics[width=0.12\textwidth]{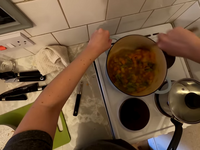}
        \includegraphics[width=0.12\textwidth]{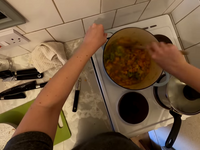}
        \includegraphics[width=0.12\textwidth]{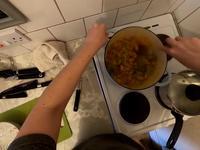} \\
        \hline

        \multirow{3}{*}{\rotatebox[origin=c]{90}{Saudi}} &
        \includegraphics[width=0.12\textwidth]{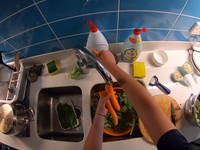}
        \includegraphics[width=0.12\textwidth]{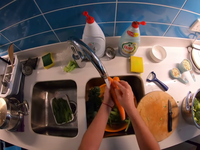}
        \includegraphics[width=0.12\textwidth]{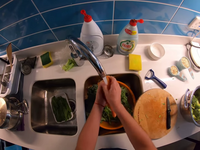}
        \includegraphics[width=0.12\textwidth]{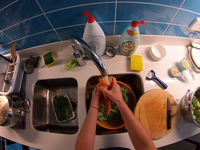}
        \includegraphics[width=0.12\textwidth]{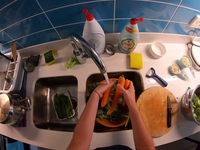}
        \includegraphics[width=0.12\textwidth]{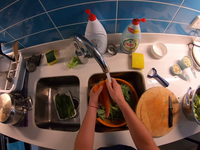} \\

        & 
        \includegraphics[width=0.12\textwidth]{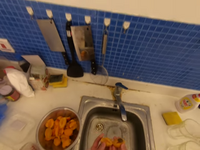}
        \includegraphics[width=0.12\textwidth]{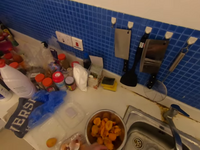}
        \includegraphics[width=0.12\textwidth]{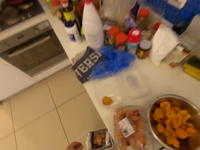}
        \includegraphics[width=0.12\textwidth]{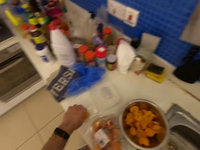}
        \includegraphics[width=0.12\textwidth]{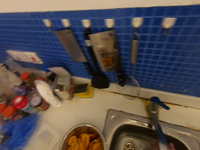}
        \includegraphics[width=0.12\textwidth]{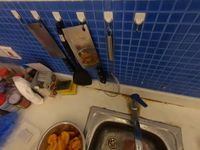} \\

        & 
        \includegraphics[width=0.12\textwidth]{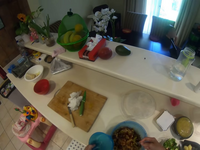}
        \includegraphics[width=0.12\textwidth]{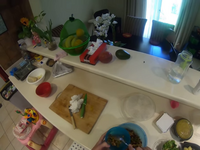}
        \includegraphics[width=0.12\textwidth]{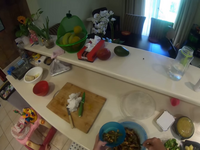}
        \includegraphics[width=0.12\textwidth]{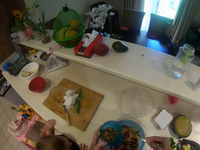}
        \includegraphics[width=0.12\textwidth]{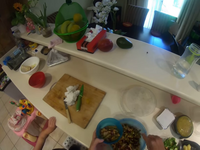}
        \includegraphics[width=0.12\textwidth]{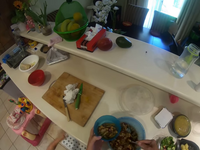} \\
        \hline

        \multirow{3}{*}{\rotatebox[origin=c]{90}{Japan}} &
        \includegraphics[width=0.12\textwidth]{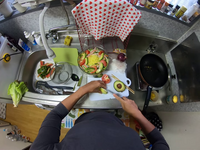}
        \includegraphics[width=0.12\textwidth]{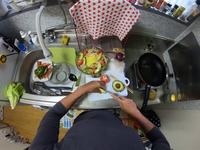}
        \includegraphics[width=0.12\textwidth]{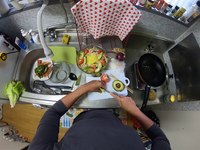}
        \includegraphics[width=0.12\textwidth]{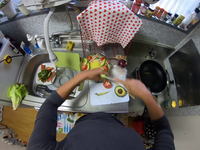}
        \includegraphics[width=0.12\textwidth]{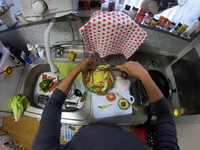}
        \includegraphics[width=0.12\textwidth]{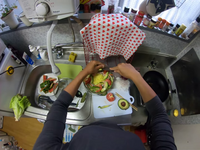} \\

        & 
        \includegraphics[width=0.12\textwidth]{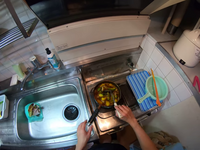}
        \includegraphics[width=0.12\textwidth]{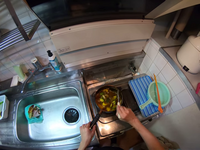}
        \includegraphics[width=0.12\textwidth]{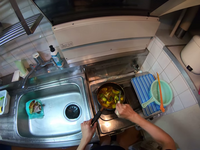}
        \includegraphics[width=0.12\textwidth]{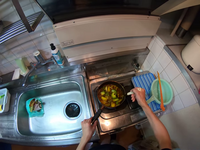}
        \includegraphics[width=0.12\textwidth]{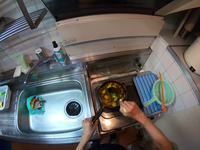}
        \includegraphics[width=0.12\textwidth]{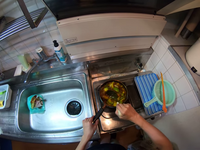} \\

        & 
        \includegraphics[width=0.12\textwidth]{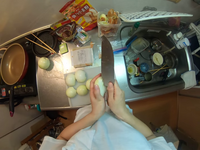}
        \includegraphics[width=0.12\textwidth]{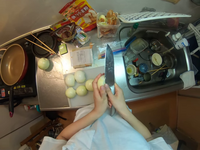}
        \includegraphics[width=0.12\textwidth]{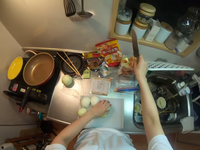}
        \includegraphics[width=0.12\textwidth]{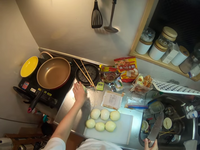}
        \includegraphics[width=0.12\textwidth]{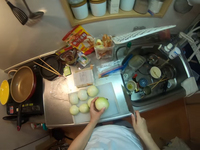}
        \includegraphics[width=0.12\textwidth]{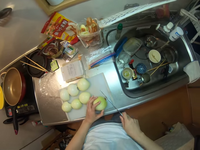} \\
        \hline
        
    \end{tabular} 
    \caption{Illustration of three representative video clips per domain from the \textit{food prep} category across five Ego4OOD domains: India, FRL, UK, Saudi, and Japan. Within each domain, clips share characteristic visual features. India kitchens differ from other domains in appearance of the walls, utensils, and overall layout. FRL consists of indoor office kitchens with distinctive background objects, and bright, consistent illumination. The UK, Saudi, and Japan domains, on the other hand, share similar kitchen utensils, layouts, and lighting. Across domains, differences in walls, objects, layouts, and lighting highlight how visual features shift while the food prep category remains the same. Horizontal lines separate domains for clarity, and each row shows frames sampled from a single video clip.}
    
    \label{fig:video_frames_domains} 
    
\end{figure*}
\begin{table}[ht!]
\centering
\caption{Ego4OOD domain shift scores as a quantitative proxy to measure covariate shift. Higher values indicate greater distributional divergence, corresponding to more challenging domains. Domains are sorted in descending order of shift score. Higher shift scores are consistently associated with reduced out-of-distribution top-1 accuracy.}
\label{tab:d_ood_scores_with_ood}
\scriptsize
\setlength{\tabcolsep}{6pt}
\begin{tabular}{lcc}
\toprule
\textbf{Domain} & $\boldsymbol{\mu \pm \sigma}$ & \textbf{Domain Shift Score} \\
\midrule
India         & \textbf{6.30} $\pm$ \textbf{0.24} & \textbf{6.78} \\
FRL        & \textbf{1.72} $\pm$ \textbf{2.09} & \textbf{5.90} \\
US-Minnesota  & \textbf{1.45} $\pm$ \textbf{2.05} & \textbf{5.55} \\
UK            & 1.41 $\pm$ 1.96                   & 5.33          \\
Saudi Arabia  & 1.34 $\pm$ 1.99                   & 5.32          \\
US-CMU        & 1.41 $\pm$ 1.95                   & 5.31          \\
Italy         & 1.34 $\pm$ 1.98                   & 5.30          \\
Japan         & 1.53 $\pm$ 1.86                   & 5.25          \\
\bottomrule
\end{tabular}
\end{table}

Figure \ref{fig:video_frames_domains} shows sample video clips from different Ego4OOD domains, while Table \ref{tab:d_ood_scores_with_ood} quantifies their differences using covariate shift scores. The highest scores are observed for India and FRL, indicating that these domains are the most visually distinct: India kitchens stand out due to differences in walls, utensils, and overall layout, whereas FRL video clips are characterized by their distinct background objects from the office, and bright, consistent indoor lighting. In contrast, Japan, Saudi, and UK have similarly low scores, reflecting their visual similarity in terms of utensils, layouts, and general appearance. This alignment between qualitative visual observations and quantitative covariate shift scores highlights how higher scores correspond to greater divergence in domain-specific visual features, while lower scores indicate domains that are more alike.

In the following section, we present our method and experiments, empirically showing a negative correlation between the proposed metric and domain-level recognition accuracy: domains with higher domain shift scores achieve lower performance, indicating that the metric provides a simple and effective predictor of domain difficulty, a finding further supported by qualitative analyses.

\section{Approach}
\subsubsection{Problem Formulation.}
We consider a domain as \( \mathcal{D} = \{\mathcal{X}, \mathcal{Y}, p(x, y)\} \), where:
\(\mathcal{X}\) is the feature space, with \(x \in \mathcal{X} \subseteq {R}^d \), representing the set of input features.
\(\mathcal{Y}\) is the label space, where \( y \in \mathcal{Y}=\{1, \ldots, K\} \) denotes the possible output categories.
\( p(x, y) \) is the joint probability distribution over feature and label spaces \( \mathcal{X} \times \mathcal{Y} \).
We are given a source domain \( \mathcal{D}_S \) and a target domain \( \mathcal{D}_T \). The source dataset is \( \mathcal{S} = \{(x_i, y_i)\}_{i=1}^{n} \) with \( n \) samples, and the target dataset is \( \mathcal{T} = \{(z_i, u_i)\}_{i=1}^{m} \) with \( m \) samples, our goal is to learn a function \( f: \mathcal{X} \to \mathcal{Y} \) from the source data that performs well on the target dataset \( \mathcal{T} \).
Given the joint probability distribution \( p(x, y) = p(x) p(y \mid x) \), we investigate the domain gap between the source and target domains. Specifically, we study the scenario where the distributional discrepancy between the source and target domains arises from differences in the feature distributions, i.e., \( p_{\mathcal{S}}(x) \neq p_{\mathcal{T}}(x) \), while the conditional probability $p(y\mid x)$ remains the same. Domain generalization involves multiple source domains \( \mathcal{D}_S^i \), where \( i \in \{1, \ldots, M\} \), each representing a distinct distribution. This setup exposes the model to a diverse range of domain distributions during training. The objective of domain generalization is to train a model on these \( M \) source domains so that it can effectively generalize to an unseen test dataset from a target domain \( \mathcal{D}_T \) \cite{wang_generalizing_2022}.

\subsubsection{Model Architecture.}
We propose {MLP-Lite}, a lightweight two-layer perceptron. We use pre-extracted SlowFast \cite{feichtenhofer2019slowfast} features. Each input consists of $T=3$ temporal features sampled from the beginning, middle, and end of the video clip, each of dimension $D=2304$, which are flattened to form the input vector. 
The MLP first projects the input to a 4096-dimensional hidden layer with layer normalization, ReLU activation, and dropout, followed by a second fully-connected layer reducing the representation to 512 dimensions with the same normalization and activation. 

\subsubsection{One-vs-all Objective.}
We adopt a \emph{one-vs-all} training strategy \cite{yang2023not} where we train $K$ independent binary classifiers--instead of a standard multi-class classifier.
Let $(\mathbf{x}_i, y_i)$ denote a training sample and $y_i \in \{1,\dots,K\}$ its ground-truth class label. We construct a one-hot target vector where $y_i^{(k)} = 1$ if the sample belongs to class $k$ and $0$ otherwise. Let $\hat{y}_i^{(k)} \in [0,1]$ be the predicted probability that sample $i$ belongs to class $k$.  
The binary cross-entropy loss for the $k$-th classifier of a sample $i$ is then given by:
\begin{equation}
\mathcal{L}_i^{(k)} = - y_i^{(k)} \log \hat{y}_i^{(k)} - (1 - y_i^{(k)}) \log (1 - \hat{y}_i^{(k)}),
\end{equation}
Each of the $K$ classifiers is trained to predict whether a sample belongs to class $k$ or not. Treating classes independently allows the model to learn distinct decision boundaries, reducing interference between visually similar classes under covariate shift \cite{padhy2020revisiting,cheng2024unified,saito2021ovanet}. 
\begin{table*}[t!]
\centering
\caption{Top-1 accuracy (\%) on ARGO1M. Best results in \textbf{bold}.}
\label{tab:argom1}
\resizebox{\textwidth}{!}{%
\setlength{\tabcolsep}{1pt}
\begin{tabular}{@{}lccccccccccc|c@{}}
\toprule
\textbf{Method} 
& \textbf{txt}
& \textbf{Ga-US} 
& \textbf{Cl-US} 
& \textbf{Kn-IND} 
& \textbf{Sh-IND} 
& \textbf{Bu-US} 
& \textbf{Me-SAU} 
& \textbf{Sp-US} 
& \textbf{Co-JPN} 
& \textbf{Ar-ITA} 
& \textbf{Pl-US}
& \textbf{Avg} \\
\midrule

Random
& -- 
& 8.00 & 10.64 & 9.13 & 14.36 & 9.55 & 13.04 & 8.35 & 10.13 & 9.86 & 15.68 & 11.58 \\

ERM
& $\times$
& 20.75 & 22.35 & 18.69 & 22.14 & 20.73 & 23.51 & 18.97 & 24.81 & 22.75 & 23.29 & 21.76 \\

\midrule
CORAL
& $\times$
& 22.14 & 22.55 & 19.07 & 24.01 & 22.18 & 24.31 & 19.16 & 25.36 & 23.89 & 25.96 & 22.87 \\

DANN
& $\times$
& 22.42 & 23.85 & 19.27 & 22.89 & 22.23 & 23.70 & 18.64 & 25.86 & 23.86 & 23.28 & 22.70 \\

MADA
& $\times$
& 22.42 & 23.60 & 19.66 & 24.46 & 22.08 & 24.64 & 19.59 & 25.87 & 23.84 & 24.78 & 23.19 \\

Mixup
& $\times$
& 21.97 & 22.21 & 19.90 & 23.81 & 21.45 & 24.35 & 19.01 & 25.90 & 23.85 & 24.41 & 22.89 \\

BoDA
& $\times$
& 22.17 & 22.78 & 19.62 & 22.94 & 21.46 & 23.97 & 19.18 & 25.68 & 23.92 & 24.90 & 22.86 \\

DocPrompt
& \checkmark
& 21.92 & 22.77 & 20.40 & 23.67 & 22.75 & 24.67 & 18.24 & 25.04 & 24.74 & 25.24 & 22.92 \\

\midrule
CIR \cite{plizzari2023can}
& \checkmark
& 23.39 & 24.52 & 21.02 & 26.62 & 24.64 & \textbf{27.00} & \textbf{19.66} & 25.42 & 25.71 & 30.17 & \textbf{24.92} \\

EgoZAR \cite{peirone2025egocentric}
& \checkmark
& \textbf{24.53} & \textbf{26.12} & \textbf{21.70} & 25.82 & 24.05 & 24.88 & 18.91 & \textbf{26.02} & \textbf{26.05} & 29.94 & 24.90 \\

\midrule

MLP-Lite (ours)
& $\times$
& 22.53 & 24.29 & 20.72 & \textbf{27.14} & \textbf{24.98} & 26.01 & 18.36 & 25.13 & 24.39 & \textbf{30.33} & 24.38 \\

\bottomrule
\end{tabular}%
} 
\end{table*}
\section{Experiments}
We evaluate our method on the Argo1M and Ego4OOD benchmarks, comparing against standard baselines and state-of-the-art methods, including CIR \cite{plizzari2023can} and EgoZAR \cite{peirone2025egocentric}.

\subsubsection{Implementation Details.}
We use pre-extracted SlowFast features provided by Ego4D, generated with a SlowFast 8$\times$8 network and a ResNet-101 backbone trained on Kinetics-400. Each feature vector corresponds to a temporal window of 32 frames, extracted with a stride of 16 frames. These features serve as input to MLP-Lite for all of our experiments. Models are trained using Adam with a learning rate of 0.01, a batch size of 128, and dropout of 0.9 on hidden layers. Training is performed for 100 epochs on a single GPU with a fixed random seed. 
For comparison, we adopt the CIR model using the training configuration from \cite{plizzari2023can}, with a batch size of 128, 50 epochs, and the Adam optimizer with a learning rate of 0.0002.

\subsubsection{Argo1M Results.}Table~\ref{tab:argom1} reports top-1 accuracy on the Argo1M benchmark. For a consistent comparison, we adopt the baselines and the CIR model configuration from \cite{plizzari2023can}. {MLP-Lite} is evaluated against these standard domain generalization baselines as well as state-of-the-art methods that leverage textual annotations. Despite its simplicity and smaller parameter count, {MLP-Lite} achieves a competitive average of 24.38\%, outperforming ERM and all other DG baselines. On several domains, it even surpasses the SOTA, while remaining competitive on the others.

\begin{table*}[ht]
\centering
\caption{Model size against average top-1 accuracy on Argo1M.}
\begin{tabular}{lccc}
\toprule
\textbf{Method} & \textbf{top-1 acc.(\%)} & \textbf{trainable params(M)} & \textbf{total params(M)} \\
\midrule
DANN      & 22.87 & 30.96 & 30.96 \\
DoPrompt  & 22.92 & 33.23 & 33.23 \\
CIR       & 24.92 & 31.11 & 182.88 \\
MLP-Lite (ours)      & 24.38 & 30.42 & 30.42 \\
\bottomrule
\end{tabular}
\end{table*}

\subsubsection{Ego4OOD Results.}
Table~\ref{tab:ego4ood_results} reports Top-1 accuracy on Ego4OOD across eight domains. {MLP-Lite} achieves a competitive average of 54.02\%, closely matching CIR (55.68\%) while surpassing it on several domains, including Japan, FRL, US-CMU, and UK. Most notable is the FRL domain, where {MLP-Lite} gains substantial improvement over CIR (36.16\% vs.\ 25.74\%).
The out-of-distribution experiments further confirm that domains with higher domain shift scores (see Table \ref{tab:d_ood_scores_with_ood}) are more challenging, exhibiting lower Top-1 accuracy. This demonstrates that the proposed metric provides a simple and effective predictor of domain difficulty.

\begin{table*}[ht!]
    \caption{Top-1 accuracy (\%) on Ego4OOD. Best results per column are in \textbf{bold}.}
    \centering
    \resizebox{\textwidth}{!}{%
    \begin{tabular}{lclcrrccc|c}
        \toprule
        Method
        & US-Minn.
        & Japan
        & FRL
        & Saudi
        & Italy
        & US-CMU
        & UK
        & India
        & Avg. \\
        \midrule

        Random
        & 16.13 & 15.30 & 10.70 & 14.32 & 12.93 & 11.28 & 14.22 & 12.91 & 13.47 \\
        \hline
        CIR \cite{plizzari2023can}
        & \textbf{50.34} & 77.34 & {25.74} & \textbf{77.73} & \textbf{52.73} & {51.86} & 63.28 & \textbf{46.40} & \textbf{55.68} \\
        
        MLP-Lite (ours)
        & 49.47 & \textbf{77.73} & \textbf{36.16} & 53.55 & 51.95 & \textbf{52.12} & \textbf{65.36} & 45.83 & 54.02  \\

        \bottomrule
    \end{tabular}}
    \label{tab:ego4ood_results}
\end{table*}

\section{Conclusion}

We demonstrate that a simple two-layer network trained with a one-vs-all objective performs on par with state-of-the-art out-of-distribution generalization models on egocentric video data, while using fewer parameters and without relying on the textual modality leveraged by other approaches. This highlights that careful network tuning can achieve competitive performance. 

Furthermore, we introduce {Ego4OOD}, a benchmark curated from the Ego4D dataset with high-quality moment-level annotations and semantically coherent categories. By reducing label ambiguity and clearly defining concept boundaries, Ego4OOD mitigates concept shift while preserving covariate diversity across eight distinct domains, providing a more reliable and interpretable setting for evaluating domain generalization methods.

In addition, we propose a domain shift scoring metric as a quantitative proxy for domain difficulty. Domains with higher shift scores exhibit stronger covariate shifts, consistent with the qualitative analyses presented in Fig.~\ref{fig:video_frames_domains}. 
Our clustering-based domain shift score offers a practical measure of covariate shift in egocentric video, allowing us to quantify domain difficulty directly in feature space. Compared to the NICO++ covariate metric \cite{zhang2023nico++}, which is theoretically grounded and provides formal guarantees, our method is easier to compute and scales to high-dimensional video features. NICO++ also introduces a metric for concept shift, based on differences in labeling functions across domains. Our current analysis on concept shift merely relies on qualitative examples, pointing to a clear direction for future work on developing robust, quantitative measures of concept shift to complement covariate assessments and further improve the evaluation of domain generalization benchmarks for egocentric videos.


\begin{thebibliography}{99}

\bibitem{dataset_bias}
Torralba, A., Efros, A.A.: Unbiased look at dataset bias.
In: Proceedings of the IEEE Conference on Computer Vision and Pattern Recognition (CVPR), pp.~1521--1528 (2011)

\bibitem{pretrain_aug_ds}
Hendrycks, D., Mu, N., Cubuk, E.D., Zoph, B., Gilmer, J., Lakshminarayanan, B.:
AugMix: A simple data processing method to improve robustness and uncertainty.
In: International Conference on Learning Representations (ICLR) (2020)

\bibitem{HendrycksBMKWDD21}
Hendrycks, D., Basart, S., Mu, N., Kadavath, S., Wang, F., Dorundo, E., Desai, R., Steinhardt, J., Gilmer, J.:
The many faces of robustness: A critical analysis of out-of-distribution generalization.
In: Proceedings of the IEEE/CVF International Conference on Computer Vision (ICCV), pp.~8340--8349 (2021)

\bibitem{nat_ds_images}
Taori, R., Dave, A., Shankar, V., Carlini, N., Recht, B., Schmidt, L.:
Measuring robustness to natural distribution shifts in image classification.
In: Advances in Neural Information Processing Systems (NeurIPS) (2020)

\bibitem{syn_vs_nat}
Shankar, V., Dave, A., Roelofs, R., Ramanan, D., Recht, B., Schmidt, L.:
A systematic framework for natural perturbations from videos.
arXiv preprint arXiv:1906.02168 (2019)

\bibitem{augmix}
Cubuk, E.D., Zoph, B., Mane, D., Vasudevan, V., Le, Q.V.:
AutoAugment: Learning augmentation strategies from data.
In: Proceedings of the IEEE/CVF Conference on Computer Vision and Pattern Recognition (CVPR), pp.~113--123 (2019)

\bibitem{PACS}
Li, D., Yang, Y., Song, Y.-Z., Hospedales, T.M.:
Deeper, broader and artier domain generalization.
In: Proceedings of the IEEE International Conference on Computer Vision (ICCV), pp.~5542--5550 (2017)

\bibitem{office_home}
Venkateswara, H., Eusebio, J., Chakraborty, S., Panchanathan, S.:
Deep hashing network for unsupervised domain adaptation.
In: Proceedings of the IEEE Conference on Computer Vision and Pattern Recognition (CVPR) (2017)

\bibitem{domainnet}
Peng, X., Bai, Q., Xia, X., Huang, Z., Saenko, K., Wang, B.:
Moment matching for multi-source domain adaptation.
In: Proceedings of the IEEE/CVF International Conference on Computer Vision (ICCV) (2019)

\bibitem{coloredmnist}
Arjovsky, M., Bottou, L., Gulrajani, I., Lopez-Paz, D.:
Invariant risk minimization.
arXiv preprint arXiv:1907.02893 (2019)

\bibitem{rotatedmnist}
Ghifary, M., Kleijn, W.B., Zhang, M., Balduzzi, D.:
Domain generalization for object recognition with multi-task autoencoders.
In: Proceedings of the IEEE International Conference on Computer Vision (ICCV), pp.~2551--2559 (2015)

\bibitem{TerraIncognita}
Beery, S., Van Horn, G., Perona, P.:
Recognition in Terra Incognita.
In: Proceedings of the European Conference on Computer Vision (ECCV), pp.~456--473 (2018)

\bibitem{epickitchens}
Damen, D., Doughty, H., Farinella, G.M., Fidler, S., Furnari, A., Kazakos, E., Moltisanti, D., Munro, J., Perrett, T., Price, W., et al.:
Scaling egocentric vision: The EPIC-Kitchens dataset.
In: Proceedings of the European Conference on Computer Vision (ECCV), pp.~720--736 (2018)

\bibitem{grauman_ego4d_nodate}
Grauman, K., et al.:
Ego4D: Around the world in 3,000 hours of egocentric video.
In: Proceedings of the IEEE/CVF Conference on Computer Vision and Pattern Recognition (CVPR), pp.~18995--19012 (2022)

\bibitem{charadesego}
Sigurdsson, G.A., Gupta, A., Schmid, C., Farhadi, A., Alahari, K.:
Charades-Ego: A large-scale dataset of paired third- and first-person videos.
arXiv preprint arXiv:1804.09626 (2018)

\bibitem{wang2013action}
Wang, H., Schmid, C.:
Action recognition with improved trajectories.
In: Proceedings of the IEEE International Conference on Computer Vision (ICCV), pp.~3551--3558 (2013)

\bibitem{zhao2018trajectory}
Zhao, Y., Xiong, Y., Lin, D.:
Trajectory convolution for action recognition.
In: Advances in Neural Information Processing Systems (NeurIPS), vol.~31 (2018)

\bibitem{peirone2025egocentric}
Peirone, S.A., Goletto, G., Planamente, M., Bottino, A., Caputo, B., Averta, G.:
Egocentric zone-aware action recognition across environments.
Pattern Recognition Letters \textbf{188}, 140--147 (2025)

\bibitem{nagarajan2020ego}
Nagarajan, T., Li, Y., Feichtenhofer, C., Grauman, K.:
Ego-Topo: Environment affordances from egocentric video.
In: Proceedings of the IEEE/CVF Conference on Computer Vision and Pattern Recognition (CVPR), pp.~163--172 (2020)

\bibitem{plizzari2023can}
Plizzari, C., Perrett, T., Caputo, B., Damen, D.:
What can a cook in Italy teach a mechanic in India? Action recognition generalisation over scenarios and locations.
In: Proceedings of the IEEE/CVF International Conference on Computer Vision (ICCV), pp.~13656--13666 (2023)

\bibitem{farahani_brief_2020}
Farahani, A., Voghoei, S., Rasheed, K., Arabnia, H.R.:
A brief review of domain adaptation.
In: Advances in Data Science and Information Engineering, pp.~877--894. Springer (2020)

\bibitem{wang_generalizing_2022}
Wang, J., Lan, C., Liu, C., Ouyang, Y., Qin, T., Lu, W., Chen, Y., Zeng, W., Yu, P.S.:
Generalizing to unseen domains: A survey on domain generalization.
IEEE Transactions on Knowledge and Data Engineering (2022)

\bibitem{tsipras2020imagenet}
Tsipras, D., Santurkar, S., Engstrom, L., Ilyas, A., Madry, A.: 
From ImageNet to image classification: Contextualizing progress on benchmarks. 
In: International Conference on Machine Learning (ICML), pp. 9625--9635 (2020)

\bibitem{zhang2023nico++}
Zhang, X., He, Y., Xu, R., Yu, H., Shen, Z., Cui, P.: 
Nico++: Towards better benchmarking for domain generalization. 
In: Proceedings of the IEEE/CVF Conference on Computer Vision and Pattern Recognition (CVPR), pp. 16036--16047 (2023)

\bibitem{quionero2009dataset}
Quionero-Candela, J., Sugiyama, M., Schwaighofer, A., Lawrence, N.D.: 
Dataset shift in machine learning. 
The MIT Press (2009)

\bibitem{feichtenhofer2019slowfast}
Feichtenhofer, C., Fan, H., Malik, J., He, K.: 
SlowFast networks for video recognition. 
In: Proceedings of the IEEE/CVF International Conference on Computer Vision (ICCV), pp. 6202--6211 (2019)

\bibitem{padhy2020revisiting}
Padhy, S., Nado, Z., Ren, J., Liu, J., Snoek, J., Lakshminarayanan, B.: 
Revisiting one-vs-all classifiers for predictive uncertainty and out-of-distribution detection in neural networks. 
arXiv preprint arXiv:2007.05134 (2020)

\bibitem{yang2023not}
Yang, Y., Zhang, Y., Song, X., Xu, Y.: 
Not all out-of-distribution data are harmful to open-set active learning. 
Advances in Neural Information Processing Systems, 36, 13802--13818 (2023)

\bibitem{cheng2024unified}
Cheng, Z., Zhang, X.-Y., Liu, C.-L.: 
Unified classification and rejection: A one-versus-all framework. 
Machine Intelligence Research, 21(5), 870--887 (2024)

\bibitem{saito2021ovanet}
Saito, K., Saenko, K.: 
Ovanet: One-vs-all network for universal domain adaptation. 
In: Proceedings of the IEEE/CVF International Conference on Computer Vision, pp. 9000--9009 (2021)

\end{thebibliography}
\end{document}